\documentclass[journal]{IEEEtran}

\usepackage{amsmath}
\usepackage{graphicx}
\usepackage{xcolor}
\usepackage{amssymb}
\usepackage{subcaption}
\usepackage{array}
\usepackage{tabularx}
\usepackage{balance}
\usepackage[all]{comment} 

\DeclareMathOperator*{\minimize}{minimize}
\ifCLASSINFOpdf
\else
\fi
\begin{document}
%
\title{Efficient and Compliant Control Framework for Versatile Human-Humanoid Collaborative Transportation}
%
%
%

\author{Shubham~S.~Kumbhar,
       Abhijeet~M.~Kulkarni,
       and~Panagiotis~Artemiadis$^*$, \textit{IEEE Senior Member}


\thanks{Shubham S. Kumbhar, Abhijeet M. Kulkarni, and Panagiotis Artemiadis are with the Mechanical Engineering Department, at the University of Delaware, Newark, DE 19716, USA. {\tt\small shubhamk@udel.edu, amkulk@udel.edu, partem@udel.edu}}
\thanks{$^*$Corrresponding author}}%


%
%

\markboth{}{}
%



\maketitle

\begin{abstract}

We present a control framework that enables humanoid robots to perform collaborative transportation tasks with a human partner. The framework supports both translational and rotational motions, which are fundamental to co-transport scenarios. It comprises three components: a high-level planner, a low-level controller, and a stiffness modulation mechanism. At the planning level, we introduce the Interaction Linear Inverted Pendulum (I-LIP), which, combined with an admittance model and an MPC formulation, generates dynamically feasible footstep plans. These are executed by a QP-based whole-body controller that accounts for the coupled humanoid–object dynamics. Stiffness modulation regulates robot–object interaction, ensuring convergence to the desired relative configuration defined by the distance between the object and the robot’s center of mass. We validate the effectiveness of the framework through real-world experiments conducted on the Digit humanoid platform. To quantify collaboration quality, we propose an efficiency metric that captures both task performance and inter-agent coordination. We show that this metric highlights the role of compliance in collaborative tasks and offers insights into desirable trajectory characteristics across both high- and low-level control layers. Finally, we showcase experimental results on collaborative behaviors, including translation, turning, and combined motions such as semi-circular trajectories, representative of naturally occurring co-transportation tasks.
\end{abstract}


%
\IEEEpeerreviewmaketitle

\section{INTRODUCTION}
\IEEEPARstart{H}{umans} routinely collaborate to accomplish physically demanding tasks, such as moving furniture across a room, carrying packages in a warehouse, or assisting elderly individuals with mobility. These activities underscore the importance of coordination, adaptability, and collaborative effort in achieving common objectives. Extending such collaboration to robots has the potential to transform multiple domains, from logistics and manufacturing to healthcare and home assistance, by enabling humans and robots to work together as integrated partners. Over the past two decades, significant research has been conducted in the field of \textit{physical human–robot interaction} (pHRI), as comprehensively reviewed in \cite{Burdet2012Framework, lasota2017survey, ajoudani2018progress, vianello2021human}.  

Physical human–robot interaction (pHRI) describes scenarios in which a human and a robot physically interact to achieve a shared objective. These interactions may be \textit{direct}, such as guiding a robot by the hand \cite{lefevre2024humanoid}, or \textit{indirect}, such as jointly manipulating an object \cite{agravante2019human}. In both cases, the partners must regulate their interaction forces to accomplish task-specific objectives. Effective pHRI requires the robot to remain dynamically stable under external loads, adapt in real-time to unknown human intent, and maintain task efficiency and safety in uncertain environments. 
These requirements make pHRI inherently more challenging than conventional robotic control, as the robot must both predict motion and execute whole-body actions while accounting for the presence of a collaborator.

\subsection{Previous Works}

Robotic manipulators have been extensively employed in pHRI for tasks such as handovers \cite{Huber2008Human,Edsinger2007Human}, carrying unknown loads \cite{MUJICA2023Robust}, and collaborative lifting \cite{Rus2019Sharing}. For instance, the authors in \cite{Rus2019Sharing} leveraged human arm muscle activity to infer the desired lifting height of an object during cooperative tasks. Similarly, the authors in  \cite{Calinon2014Learning} demonstrated that collaborative skills can be transferred from human–human demonstrations to robotic manipulators. 
These studies demonstrate the effectiveness of robotic manipulators in physical interaction tasks, but their fixed base restricts the reachable workspace and prevents the mobility needed for tasks such as collaboratively transporting objects over long distances.

To address the limitations of fixed-base manipulators, mobile manipulator setups have been proposed \cite{Kosuge2000Mobile,Kosuge2001Control,Kosuge2002Handling}. These works developed control strategies that regulate the compliance between the mobile base and the object being carried. More recently, authors in \cite{Xing2021Human} introduced a control framework based on skill transfer for collaborative lifting and carrying tasks using wheeled mobile manipulators. Beyond object transportation, a compliant control strategy has been proposed in assisted walking \cite{Li2024Compliant}. While these robots provide greater maneuverability and extend the operational workspace, these systems are fundamentally optimized for flat, structured surfaces \cite{Bruzzone2012Review}. In contrast, human environments often contain irregular terrain, stairs, and cluttered layouts, where wheeled platforms are less effective \cite{Bruzzone2012Review}. This gap highlights the need for robots with legged bases, whose inherent versatility and maneuverability make them better suited for collaboration in human-centric environments.

The efficiency and fluidity observed in human–human collaboration provide a strong motivation for extending similar principles to humanoid robots, giving rise to the field of \textit{physical human–humanoid interaction} (pHHI). Unlike wheeled mobile manipulators, humanoids can naturally navigate uneven terrains \cite{radosavovic2024learning}, climb stairs \cite{caron2019stair}, and operate within cluttered or constrained spaces \cite{Poulakakis2022Sequential, Kumbhar2025Finite}, making them particularly well-suited for tasks in warehouses, homes, and healthcare settings. 

Beyond functional capabilities, humanoids have attracted increasing attention in both industry and academia due to their dexterity and versatility in performing a wide range of tasks. Recent market reports even project widespread global deployment of humanoids by 2040 \cite{neumann2025humanoid}, underscoring their anticipated impact across multiple sectors. 
However, the task of pHHI is considerably more challenging. Unlike wheeled bases, which are mechanically simpler and dynamically more stable, the task of maintaining balance, stability, and robustness for legged humanoids is much more challenging. These challenges necessitate advanced control frameworks that can ensure safe, stable, and compliant interaction between humanoids and human partners.

One of the earliest demonstrations of pHHI was presented in \cite{Hirukawa2003Cooperative}, where the HRP-2P humanoid robot was used to collaboratively move a panel with a human operator. 
The authors in \cite{Agravante2014Collaborative} implemented a control architecture for humanoid collaboration based on a Stack of Tasks formulation. Their approach provided a vision algorithm to estimate the desired object states. These desired states were subsequently used for admittance control of the humanoid. A different approach was adopted by authors in \cite{motahar2015integrating, motahar2017steering}, where interaction forces were leveraged for adaptation of the walking limit cycle, while the arms were impedance controlled to ensure safe physical interaction. Subsequent works have explored the use of intent-carrying forces to enable switching, either among a library of limit cycles \cite{veer2017adaptation} or among Dynamic Motion Primitives (DMPs) \cite{chand2022interactive}.

Current research on humanoid walking with optimization-based methods typically formulates the control hierarchy at two levels. At the high level, center of mass (CoM) trajectories and footstep locations are planned, while at the low level, these plans are realized on the physical robot through whole-body control. Model Predictive Control (MPC) has emerged as a widely used approach for high-level planning, leveraging simplified dynamic models such as the Linear Inverted Pendulum (LIP) \cite{narkhede2023overtaking, Kajita2001LIP}, and the Single Rigid Body (SRB) \cite{Ding2022Orientation}. At the low level, Quadratic Programming (QP) formulations are employed to compute feasible joint torques and contact forces that ensure stability and adherence to the planned trajectories \cite{Nguyen2015Safety}. Current work in pHHI, including our own, adopts a similar hierarchical structure, employing an MPC-based high-level planner for footstep generation, coupled with a QP-based low-level controller to track and execute the motion plan on the Digit humanoid robot \cite{agravante2019human, Kumbhar2025MPC}.

The work in \cite{Kheddar2016Walking} introduced a hierarchical MPC–QP framework specifically for pHHI. The control for the humanoid as a follower incorporated an impedance model of the CoM directly into the MPC cost function to regulate interaction behavior, and footstep patterns were generated accordingly. However, they used a quasi-static motion assumption to transform the external wrench from the sensor frame to the CoM frame, and it was also assumed to remain constant over the MPC horizon. Furthermore, the impedance model specifies a desired impedance between the robot CoM and the origin, implying that the human partner needed to exert significant effort to displace the robot from its starting configuration. Later, in \cite{Kheddar2019Human}, these high-level plans were tracked via a QP that explicitly considered external forces. While knowledge of these forces enabled compensatory actions for stability and trajectory tracking, the robot did not actively regulate them. 

The work in \cite{Kumbhar2025MPC} addressed these challenges by introducing an admittance model that regulated the relative motion between the humanoid's CoM and the object, coupled with an object-informed QP for low-level control. 
However, the stiffness modulation in their approach relied on a global trajectory from the admittance model, causing the controller to behave rigidly when the CoM deviated, and preventing \textit{local compliance}, i.e., a flexible response to transient deviations or perturbations consistent with the desired compliance characteristics. Moreover, neither of these studies addressed turning maneuvers, which are essential for collaborative transportation in human environments.


\subsection{Main Contributions}

pHHI consists of the coupled system of the human and the humanoid. Although it is important for the humanoid to be stable while assisting with the task, the efforts of the human and the effect of the control framework on the human's efficiency during the task have often not been addressed \cite{Kumbhar2025MPC, Kheddar2019Human}. In this work, we not only focus on the humanoid's ability to carry out the task efficiently but also on the efficiency with which humans can carry out the same collaborative task. To develop such a human-oriented control framework, compliance between the human and the robot is necessary. This also helps the robot stay stable, as there is now a cushioning layer between the human and the humanoid. Authors in \cite{Kumbhar2025MPC} make the footstep planner capable of generating footstep plans that enable the robot to have a certain desired compliance. But, as discussed above, this compliance is not ``local'' in nature. Moreover, they do not explore the use of hands towards this compliance.

This paper presents a comprehensive framework for enhancing the compliant behavior of humanoid robots when physically collaborating with humans to transfer objects, with a particular focus on scenarios involving the maneuvering of these objects. We extend the work done in \cite{Kumbhar2025MPC}, to propose a three-level hierarchical control architecture that integrates: (i) an admittance model to regulate the desired interaction behavior between the humanoid and the object, including turning maneuvers arising from object rotation; (ii) a high-level planner based on the \textit{Interaction Linear Inverted Pendulum} (I-LIP), a \textbf{novel} simplified model incorporating variable-stiffness springs and dampers between the centers of mass (CoMs) of the robot and the object, formulated within an MPC framework; and (iii) a low-level QP that explicitly incorporates object dynamics to regulate interaction forces and control the object’s position and orientation.  

A key feature of our framework is that compliance arises from both footstep planning and hand-object interactions. While the MPC generates foot placements that encode compliant responses to perturbations, additional objectives on the CoMs of the robot and the object in the QP ensure that the hands contribute actively to compliance. To ensure convergence to a desired distance between the object and the humanoid, we implement a gain modulation strategy for adjusting the spring constant of the I-LIP. These features promote \emph{local} compliance properties to the humanoid.

Additionally, we conduct an extensive analysis of the resulting framework to evaluate the role of compliance in enhancing human efficiency during collaborative object transportation while preserving robot stability. In particular, we introduce a quantitative efficiency metric to show that a low-compliance objective for footstep planning and a high-compliance objective for low-level control improve the efficiency of collaboration, while simultaneously preserving partner stability. Another distinguishing aspect of this work is its focus on carrying objects of substantial weight. The experimental validation considers a 15~kg box, representative of loads that typically require two humans to co-manipulate, a scenario not addressed in existing literature.  

We validate the proposed framework on the Digit humanoid (Agility Robotics) through real-world experiments. Our results demonstrate that the framework enables compliant human–humanoid collaboration during complex maneuvers involving both translation and turning, marking one of the first steps toward deploying humanoids for physical collaboration in human-centered workspaces while explicitly considering human efficiency and stability.

\subsection{Advancements and Extensions}

We expand the scope of collaborative behaviors by incorporating lateral translations and turning maneuvers, moving beyond the straight-line walking simulated in \cite{Kumbhar2025MPC}. We also reformulate the control objectives and stiffness modulation so that trajectories are executed with \emph{local} rather than global compliance, enabling more adaptive and robust interactions. In addition, we introduce explicit compliance objectives for the humanoid’s hands, enabling them to actively contribute to the co-transportation process, in contrast to \cite{Kumbhar2025MPC}, where compliance was achieved solely through footstep patterns. The framework is further extended to manipulate the orientation of the object, allowing the robot to achieve an intended height unknown to it while simultaneously regulating orientation, in contrast to the direct regulation of height without any orientation control in \cite{Kumbhar2025MPC}. We also propose a quantitative metric for evaluating the efficiency of the collaborative task, providing a principled basis for comparison across different settings. Finally, we present an extensive set of real-world experiments demonstrating how high- and low-compliance objectives, at both the high- and low-level control layers, influence stability, efficiency, and ergonomics, and showcasing the humanoid’s capability to perform a wide range of collaborative tasks.
\begin{figure} 
\centerline{\includegraphics[width=3.6in]{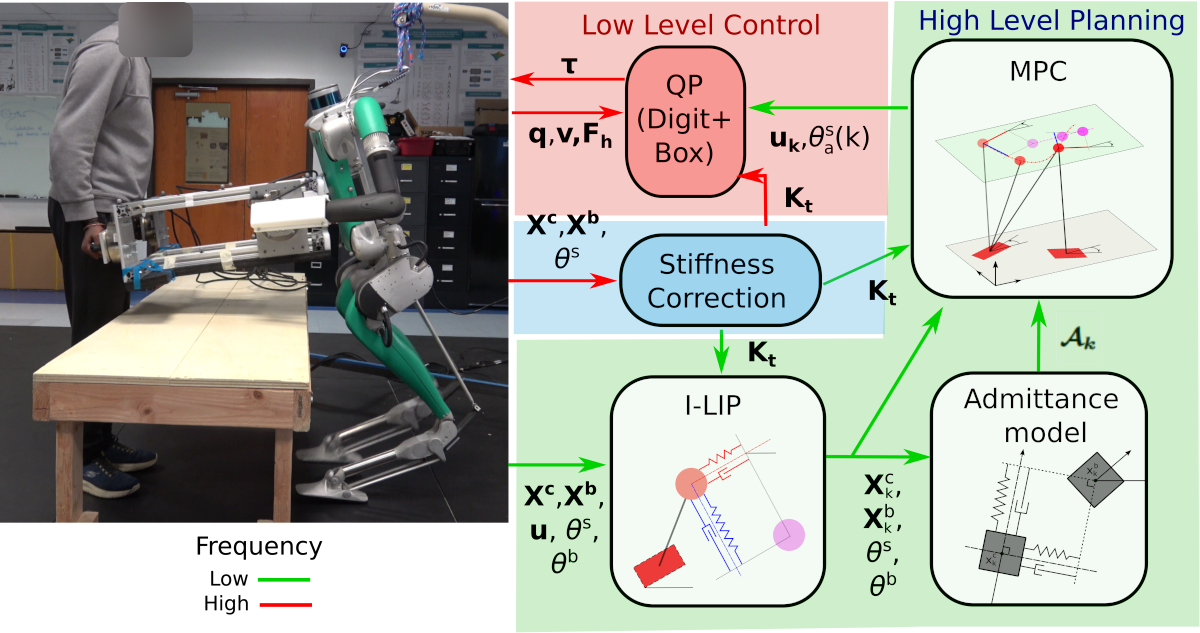}}
\caption{Proposed control framework (right) for collaborative transportation tasks between a human and a humanoid robot (left).}
\label{complete_frame_tro}
\end{figure}
\begin{figure*}
\centering
\includegraphics[scale = 0.61]{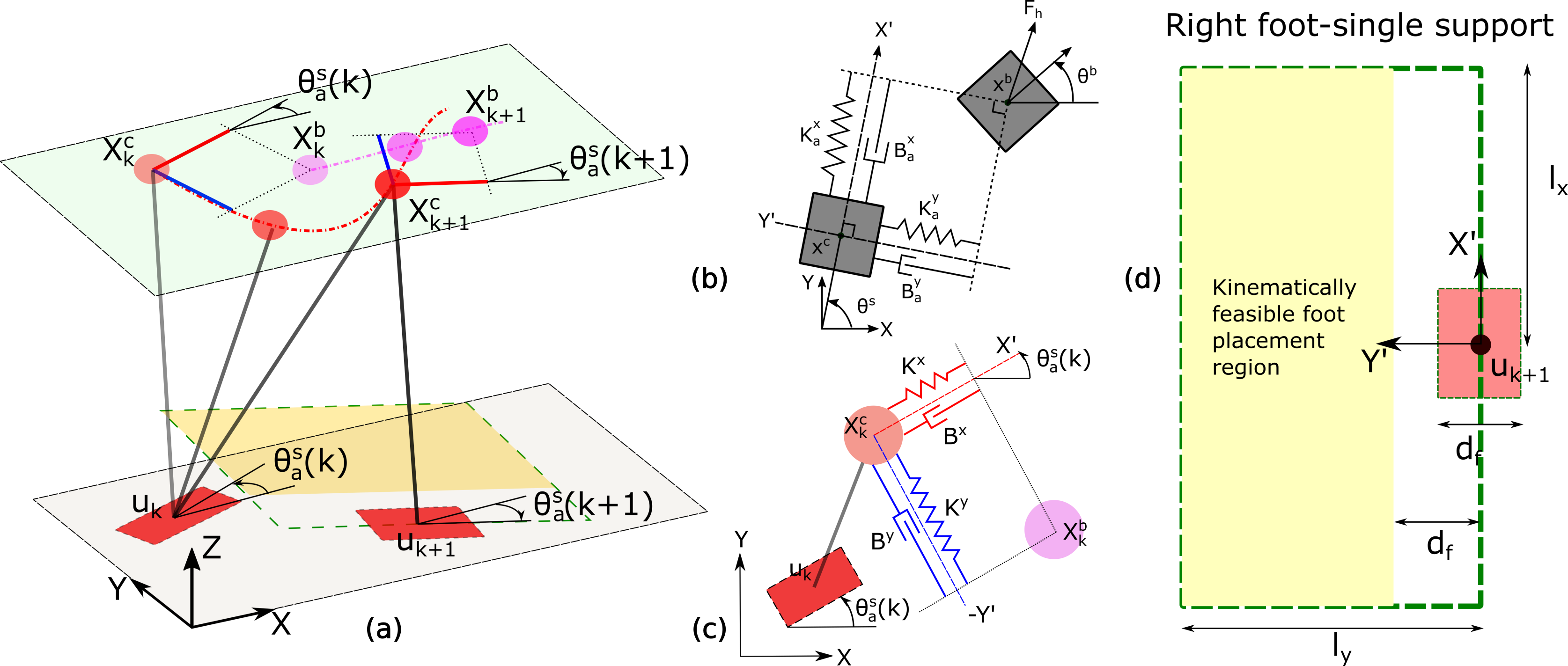}
\caption{
    \textbf{(a)} I-LIP dynamics during the $k^\mathrm{th}$ walking step with the left leg as the stance foot, denoted by $\mathbf{u}_k$. The red and blue solid lines represent spring-damper elements aligned along the local $X'$ and $Y'$ axes, respectively. The red dashed trajectory indicates the evolution of the robot's center of mass (CoM), while the purple dashed trajectory corresponds to the object's CoM. Temporal progression is illustrated through decreasing transparency, where increased opacity represents forward movement in time. The red rectangle denotes the stance foot position, and the yellow rectangle indicates the admissible region for the next foot placement based on kinematic feasibility.
    \textbf{(b)} Admittance-based interaction model, abstracted as two point masses, representing the CoMs of the robot and the object interconnected by spring damper pairs in the $X'$ and $Y'$ directions. This model captures the compliant coupling between the robot and the object and enables responsive interaction.
    \textbf{(c)} 
    Top-down view of the I-LIP configuration at the beginning of the $k^\mathrm{th}$ step, highlighting the orientation of spring-damper elements in the horizontal $X'$-$Y'$ plane. 
    \textbf{(d)} Admissible foot placement region with the right leg as the stance foot. The yellow region defines the kinematically feasible area for placing the left foot in the next step. This region is constrained by the robot's reachability and ensures both feasibility and stability in step planning.
    }
\label{all_figures}
\end{figure*}
\section{CONTROL ARCHITECTURE FOR COLLABORATIVE TRANSPORTATION}
We address collaborative transportation tasks in which the Digit humanoid and a human jointly transfer an object (box), as illustrated in Fig. \ref{complete_frame_tro}. In this setting, the human assumes the role of the leader, actively maneuvering the object according to their intent, while the humanoid functions as a follower, passively adapting its motion to the object's movement. The proposed framework enables the humanoid to exhibit a spectrum of behaviors ranging from highly compliant to more assertive motion. Building upon the work presented in \cite{Kumbhar2025MPC}, we extend the follower's capabilities to include forward, lateral, and rotational maneuvers, thereby providing the human collaborator greater freedom in executing the co-transportation task.

Figure \ref{complete_frame_tro} (right) presents the proposed control framework for physical human–humanoid collaboration, highlighting its key components: the I-LIP model, the admittance model, MPC formulation, the whole-body controller (WBC), and the stiffness modulation module. These components operate in a sequential and tightly integrated manner to enable stable, compliant, and adaptive co-transportation behavior.

The I-LIP and admittance models, together with the MPC, form the high-level planning layer, which generates footstep placements and CoM trajectories. The I-LIP illustrated in Fig. \ref{all_figures}(a) serves as a simplified predictive model for planning purposes, while the admittance model illustrated in Fig. \ref{all_figures}(b) dictates the desired compliant behavior between the humanoid and the object. By tuning the gains of the admittance model, the robot can exhibit behaviors ranging from high compliance to low compliance. The MPC integrates the I-LIP dynamics with the desired states provided by the admittance model to compute optimal footstep locations that satisfy modeling constraints.

The resulting high-level plans are executed on the robot by the WBC, formulated as a QP. Unlike conventional humanoid control approaches that consider only the robot's own dynamics, our WBC models the coupled humanoid–object system, enabling direct regulation of interaction forces. Finally, the stiffness modulation module adjusts the parameters of the I-LIP's virtual coupling in real time to improve compliance tracking and maintain a desired relative motion between the humanoid and the object.

We begin by introducing the definitions and notations used throughout the subsequent subsections. Forward and lateral motions are defined as movements of the humanoid along its local \(X'\) and \(Y'\) axes, respectively, as illustrated in Fig.~\ref{all_figures}(c), while turning motion corresponds to rotations about the \(Z\)-axis. For clarity, both the local coordinate frame and the global (world) frame are shown in the figure.

Let \(\mathbf{x}^{\mathrm{c}} = [x^{\mathrm{c}}, y^{\mathrm{c}}]^\top\) and \(\dot{\mathbf{x}}^{\mathrm{c}} = [\dot{x}^{\mathrm{c}}, \dot{y}^{\mathrm{c}}]^\top\) denote the global \(X\)- and \(Y\)-component position and velocity, respectively, of the Digit's center of mass (CoM). The global position of the stance foot is given by \(\mathbf{u} = [u^{\mathrm{x}}, u^{\mathrm{y}}]^\top\), with the \(Z\)-component of all foot strikes assumed to be zero. Similarly, the object’s global position and velocity are denoted by \(\mathbf{x}^{\mathrm{b}} = [x^{\mathrm{b}}, y^{\mathrm{b}}]^\top\) and \(\dot{\mathbf{x}}^{\mathrm{b}} = [\dot{x}^{\mathrm{b}}, \dot{y}^{\mathrm{b}}]^\top\), respectively.

For compactness, we define the concatenated state vectors \(\mathbf{X}^\mathrm{c} = [\mathbf{x}^\mathrm{c}, \dot{\mathbf{x}}^\mathrm{c}]^\top\) for the humanoid and \(\mathbf{X}^\mathrm{b} = [\mathbf{x}^\mathrm{b}, \dot{\mathbf{x}}^\mathrm{b}]^\top\) for the object. The yaw angles of the stance foot and the object are denoted by \(\theta^{\mathrm{s}}\) and \(\theta^{\mathrm{b}}\), respectively. As shown in Fig.~\ref{complete_frame_tro}, the high-level planning block takes the tuple $(\mathbf{X}^\mathrm{c}, \mathbf{X}^\mathrm{b}, \mathbf{u}, \theta^{\mathrm{s}}, \theta^\mathrm{b})$
as input for subsequent computations in the I-LIP model, the admittance model, and the MPC module, described in the following subsections.

\subsection{Interaction Linear Inverted Pendulum (I-LIP)}
Simplified models are widely employed in a hierarchical fashion, together with WBCs, for CoM motion and footstep planning in humanoid robots \cite{Wensing2024Optimization}. This preference arises from the fact that motion planning using full-order models is computationally expensive due to the high dimensionality of the variable space, as well as the inherent nonlinearity and non-convexity of the resulting optimization problem. In this work, we also adopt a simplified model for motion planning and prediction, which is presented in detail in this section.

Figure \ref{all_figures}(a) illustrates the I-LIP model used for predicting the evolution of the robot's CoM. The I-LIP extends the classical LIP model~\cite{Kajita2001LIP} to incorporate the influence of interaction forces and object motion into footstep and CoM planning.
The I-LIP consists of a massless prismatic leg and two point masses representing the CoMs of the robot and the object. One end of the leg pivots at the stance foot contact point on the ground, while the other is attached to the point mass corresponding to the robot’s CoM. The two point masses are connected by linear springs and dampers in both the $X'$ and $Y'$ directions, as shown in Fig.~\ref{all_figures}(c). The stance foot also has a yaw (heading) angle, and the springs and dampers are oriented along the local coordinate axes defined by this heading.
In the model, the object is assumed to move with a constant velocity, denoted by 
$\mathbf{v}^\mathrm{b}_\mathrm{d}=[v^\mathrm{bx}_\mathrm{d},v^\mathrm{by}_\mathrm{d}]^\top$.
This velocity is estimated using an exponential moving average of the current measured object velocity, computed as:
\begin{align}
    \mathbf{v}_\mathrm{d}^\mathrm{b} &= \alpha \mathbf{v}_\mathrm{d}^\mathrm{b} + (1-\alpha){\mathbf{\dot{x}}}^\mathrm{b}
\end{align}
where \(\alpha \in (0,1]\) is the smoothing factor.
The \(k^\mathrm{th}\) step of the I-LIP occurs over the time interval \([kT, (k+1)T)\) between two consecutive foot strikes, where \(T\) denotes the step duration. In Fig.~\ref{all_figures}(a,c), \(\mathbf{u}_\mathrm{k} = \mathbf{u}(t = kT)\) and \(\theta^\mathrm{s}_\mathrm{k}\) represent the position and the heading (yaw) of the stance foot during the \(k^\mathrm{th}\) step. The vectors \(\mathbf{X}_\mathrm{k}^\mathrm{c} = \mathbf{X}^\mathrm{c}(t = kT)\) and \(\mathbf{X}_\mathrm{k}^\mathrm{b} = \mathbf{X}^\mathrm{b}(t = kT)\) denote the CoM state vectors of the robot (\(\mathbf{X}^\mathrm{c}\)) and the object (\(\mathbf{X}^\mathrm{b}\)), respectively, at the start of the \(k^\mathrm{th}\) step. Similarly, \(\mathbf{X}_{\mathrm{k+1}}^\mathrm{c}\) and \(\mathbf{X}_{\mathrm{k+1}}^\mathrm{b}\) represent the corresponding CoM state vectors at the start of the \((k+1)^\mathrm{th}\) step. The quantities \(\mathbf{u}_{\mathrm{k+1}}\) and \(\theta^\mathrm{s}_{\mathrm{k+1}}\) denote the position and yaw of the stance foot during the \((k+1)^\mathrm{th}\) step. The stepping sequence alternates between the left and right foot; thus, if \(\mathbf{u}_\mathrm{k}\) corresponds to the left stance foot, then \(\mathbf{u}_{\mathrm{k+1}}\) corresponds to the right stance foot.


The dynamics of the CoM of the robot in I-LIP are described by
\begin{align}
   \nonumber\mathrm{m_c}\mathbf{\ddot{x}}^\mathrm{c} =&~ \mathrm{m}_\mathrm{c}\omega_\mathrm{o}^2(\mathbf{x}^c-\mathbf{u})\\\nonumber&+\mathbf{R}(\theta^\mathrm{s})\mathbf{K}_\mathrm{t}\big(\mathbf{R}(\theta^\mathrm{s})\big)^\top\left(\mathbf{x}^\mathrm{b}_\mathrm{init}+\mathbf{v}^\mathrm{b}_\mathrm{d}t-\mathbf{x}^c-\mathbf{x}^\mathrm{d}_\mathrm{m}\right)\\&+\mathbf{R}(\theta^\mathrm{s})\mathbf{B}\big(\mathbf{R}(\theta^\mathrm{s})\big)^\top\left(\mathbf{v}^\mathrm{b}_\mathrm{d}-\mathbf{\dot{x}}^\mathrm{c}\right) 
   \label{eq:ILIP_continuous}
\end{align}
where $\omega_\mathrm{o}=\sqrt{g/h}$, $g$ is the magnitude of the gravitational acceleration, $h$ is the height of the robot's CoM from the ground, $\mathbf{x}^\mathrm{b}_\mathrm{init}$ is the initial position of the CoM of the object, and $\mathbf{x}^\mathrm{d}_\mathrm{m}$ is the x-y component natural length of the spring. $\mathbf{K}_t=diag(K^\mathrm{x}_t, K^\mathrm{y})$ is the spring constant matrix at the instant the high-level planner is initiated. 
The subscript \(t\) in \(\mathbf{K}_t\) and \(K^{\mathrm{x}}_t\) indicates that these values vary with time according to the stiffness modulation block in Fig. \ref{complete_frame_tro}. However, within the high-level planning block they are fixed to the value at the instant the planner is invoked and remain constant until the planning process is completed.
$\mathbf{B}=diag(B^\mathrm{x},B^\mathrm{y})$ is the damping constant matrix. 
By applying the analytical solutions of the non-homogeneous linear differential equations in~\eqref{eq:ILIP_continuous}, the step-to-step discrete dynamics of the I-LIP $f_\mathrm{c}$ can be derived as shown below.
\begin{equation}
    \mathbf{X}^\mathrm{c}_{k+1}=f_\mathrm{c}(\mathbf{X}^\mathrm{c}_k,  \mathbf{x}^\mathrm{b}_k, \mathbf{v}^\mathrm{b}_\mathrm{d}, \mathbf{u}_k, \theta^\mathrm{s}_k, \mathbf{K}_t, T)
    \label{eq:ILIP_discrete}
\end{equation}
Similarly, the step-to-step evolution of the CoM of the object under the constant object velocity assumption is given by:
\begin{align}
\mathbf{X}^\mathrm{b}_{k+1} = f_\mathrm{b}(
    \mathbf{x}^\mathrm{b}_{k} ,\mathbf{v}^\mathrm{b}_\mathrm{d}, T) = \begin{bmatrix}
    \mathbf{x}^\mathrm{b}_{k} + \mathbf{v}^\mathrm{b}_\mathrm{d}T \\ \mathbf{v}^\mathrm{b}_\mathrm{d}
\end{bmatrix}  
\label{function_fb}
\end{align}
If we assume that the system is in the \((k-1)^\mathrm{th}\) step, the I-LIP block in Fig.~\ref{complete_frame_tro} utilizes the functions \(f_\mathrm{c}\) and \(f_\mathrm{b}\) to predict the COM states of the robot ($\mathbf{X}^\mathrm{c}_{k}$) and the object ($\mathbf{X}^\mathrm{b}_{k}$) at the end of the current step. As illustrated in Fig.~\ref{complete_frame_tro}, the tuple \((\mathbf{X}^\mathrm{c}_k, \mathbf{X}^\mathrm{b}_k, \theta^\mathrm{s}, \theta^\mathrm{b})\) is subsequently provided to the admittance model and MPC blocks for further planning.




Before introducing the admittance model, we briefly outline certain functional details of the I-LIP. Due to the robot’s physical limitations, constraints are imposed on the allowable placement of the next foot. Figure~\ref{all_figures}(d) illustrates the feasible foot placement region for the subsequent step of the I-LIP. This region is parameterized by the following quantities:  
\(l_x\): maximum footstep length in the forward direction,  
\(l_y\): maximum footstep length in the lateral direction, and  
\(d_f\): foot width. The feasible foot placement region is defined as:
\begin{equation}
\begin{aligned}
    \mathcal{U}_k(\theta^\mathrm{s}_k) = \{(u^x,u^y) \,|\, & -l_x \leq \big(\mathbf{R}_1(\theta^\mathrm{s}_k)\big)^\top\big(u^x_{k+1} - u^x_k\big) \leq l_x, \\
    & d_f \leq n\,\big(\mathbf{R}_2(\theta^\mathrm{s}_k)\big)^\top\big(u^y_{k+1} - u^y_k\big) \leq l_y \},
    \label{kin_const}
\end{aligned}
\end{equation}
where \(n = 1\) if the stance foot is the right foot, and \(n = -1\) otherwise. The subscripts \(1\) and \(2\) in \(\big(\mathbf{R}(\cdot)\big)^\top\) denote the first and second rows of the rotation matrix, respectively.  
For the lateral $Y'$ direction, an additional constraint ensures that the left and right legs do not cross. Furthermore, a clearance equal to the foot width is included to prevent collisions between the swing foot and the stance foot.

The states predicted at the end of the step are subsequently used in the admittance model to compute the desired CoM motion corresponding to a specified compliant behavior. These results are then provided to the MPC module for generating the footstep pattern.

\subsection{Compliance Shaping via Admittance Model}
Admittance control is a widely used technique in human-robot collaboration~\cite{ajoudani2018progress} for enabling safe interaction and achieving compliant behavior of the robot during physical contact. In this approach, a high-level spring–mass–damper model, which governs the compliance, generates the desired end-effector positions and velocities. These references are then tracked by a lower-level controller.  
In this work, we adopt an admittance model together with a footstep pattern generator for a similar purpose tailored to a humanoid robot. The objective is to realize compliant behavior by leveraging both the stepping pattern and the arm motions within the whole-body control framework.

The admittance model consists of two point masses connected by a spring and a damper, as illustrated in Fig.~\ref{all_figures}(b). The robot and the object are represented as point masses with masses \(\mathrm{m}_\mathrm{c}\) and \(\mathrm{m}_\mathrm{b}\), respectively. These are coupled through a pair of springs and dampers oriented along the heading direction defined by the stance foot yaw.  

The stiffness and damping parameters of these elements can be tuned to achieve the desired compliant behavior between the robot and the object. For instance, a lower stiffness value increases the robot’s compliance relative to the object, whereas a higher stiffness reduces compliance.

We assume that the object moves with a constant velocity \(\mathbf{v}^\mathrm{b}_\mathrm{d}\), as computed in the previous subsection. Under this assumption, the dynamics of the robot’s CoM in the admittance model can be expressed as:
\begin{align}
   \nonumber \mathrm{m_c}\mathbf{\ddot{x}}^c &= \mathbf{R}(\theta^\mathrm{s})\mathbf{K}_\mathrm{a}\big(\mathbf{R}(\theta^\mathrm{s})\big)^\top\big(\mathbf{x}^\mathrm{b}-\mathbf{x}^\mathrm{c}-\mathbf{x}^\mathrm{d}\big)\\\label{eq:admittance_dyn} &+ \mathbf{R}(\theta^\mathrm{s})\mathbf{B}_\mathrm{a}\big(\mathbf{R}(\theta^\mathrm{s})\big)^\top\big(\mathbf{v}^\mathrm{b}_\mathrm{d}-\mathbf{\dot{x}}^\mathrm{c}\big)\\
    \mathbf{x}^\mathrm{b} &= \mathbf{x}^\mathrm{b}_\mathrm{init}+\mathbf{v}^\mathrm{b}_\mathrm{d}t
\end{align}
Here, $\mathbf{x}^\mathrm{b}_\mathrm{init}$ is the initial position of the CoM of the object. $\mathbf{K}_\mathrm{a} = diag(K^x_\mathrm{a}, K^y_\mathrm{a}) $ and $\mathbf{B}_\mathrm{a} = diag(B^x_\mathrm{a},B^y_\mathrm{a})$ are positive definite diagonal matrices.
 The objective of the admittance model is to provide the desired CoM states and footstep orientations of the robot at the end of each future step. For a prediction horizon of \(N\) steps, the admittance model dynamics are integrated to obtain the \(N\) predicted end-of-step COM states and corresponding footstep orientations.

Let the desired orientation of the stance foot be denoted by $\theta^\mathrm{s}_\mathrm{a}$. Similar to previous definitions, $\theta^\mathrm{s}_\mathrm{a}(k)$ denotes the desired orientation of the stance foot during the $k^\mathrm{th}$ step. The orientations are computed using a rotational spring–mass admittance model, expressed as:
\begin{align}
    \ddot{\theta}^\mathrm{s}_\mathrm{a} = k^{\mathrm{\theta}}_\mathrm{P}(\theta^\mathrm{b}_\mathrm{d}-\theta^\mathrm{s}_\mathrm{a}) - k^\theta_\mathrm{D}\dot{\theta}^\mathrm{s}_\mathrm{a}
    \label{theta_dyn}
\end{align}
Here, $k^{\mathrm{\theta}}_\mathrm{P}$ and $k^{\mathrm{\theta}}_\mathrm{D}$ are positive constants that rule the compliant nature of the stance foot yaw with respect to the object yaw. We add a small damping term ($k^\theta_\mathrm{D}\dot{\theta}^\mathrm{s}_\mathrm{a}$) to stabilize the response. \(\theta^\mathrm{b}_\mathrm{d}\) denotes the estimated intended yaw of the object. This estimate is computed in a manner similar to \(\mathbf{v}^\mathrm{b}_\mathrm{d}\), as
\begin{align}
    \theta^\mathrm{b}_\mathrm{d} = \beta\theta^\mathrm{b}_\mathrm{d} + (1-\beta)\theta^\mathrm{b}
\end{align}
where \(\beta \in (0,1]\) is the smoothing factor.
The dynamics for \(\theta^\mathrm{s}_\mathrm{a}\) \eqref{theta_dyn} are derived under the assumption that \(\theta^\mathrm{b}_\mathrm{d}\) remains constant over the next \(N\) steps. These dynamics are integrated from the initial value \(\theta^\mathrm{s}_{k-1}\) to obtain $\theta^\mathrm{s}_\mathrm{a}(k+i)~\forall~ i = 0,1,2,\ldots,N-1$. These desired orientations are further utilized in the admittance model to obtain the desired CoM states.

 It is also important to note that, within each step, the stance foot orientation is fixed at \(\theta_\mathrm{k}^\mathrm{s}\). Consequently, the actual evolution of the heading follows a stepwise profile rather than the smooth trajectories given by~\eqref{theta_dyn}. Nonetheless, these profiles coincide periodically at the end of each step.

Using the analytical solution of the non-homogeneous differential equation~\eqref{eq:admittance_dyn}, along with the desired stance foot orientations obtained above, we define a function \(f_\mathrm{a}\) that maps the robot’s CoM state at the start of the \(k^\mathrm{th}\) step, \((\mathbf{x}^\mathrm{c}_k, \mathbf{\dot{x}}^\mathrm{c}_k)\), to the CoM state at the start of the \((k+1)^\mathrm{th}\) step, \((\mathbf{x}^\mathrm{c}_{k+1}, \mathbf{\dot{x}}^\mathrm{c}_{k+1})\):
\begin{align}
    \mathbf{x}^\mathrm{c}_{k+1}, \mathbf{\dot{x}}^\mathrm{c}_{k+1} 
    = f_\mathrm{a}\big(\mathbf{x}^\mathrm{c}_{k}, \mathbf{\dot{x}}^\mathrm{c}_{k}, \mathbf{x}^\mathrm{b}_{k}, \mathbf{v}^\mathrm{b}_\mathrm{d}, \theta^\mathrm{s}_\mathrm{a}(k), T\big).
    \label{ad_dyn_discrete}
\end{align}

Using \(\mathbf{X}^\mathrm{c}_k\) and \(\mathbf{X}^\mathrm{b}_k\) as inputs (as shown in Fig.~\ref{complete_frame_tro}), and applying~\eqref{ad_dyn_discrete}, the admittance model block computes $\mathbf{x}^\mathrm{c}_\mathrm{a}(k+i) = \mathbf{x}^\mathrm{c}_{k+i},$ $ 
\mathbf{\dot{x}}^\mathrm{c}_\mathrm{a}(k+i) = \mathbf{\dot{x}}^\mathrm{c}_{k+i}$, $\forall~ i = 1, 2, \ldots, N$, where $N$ denotes the prediction horizon of the MPC, i.e., the number of future steps considered.  

Let the output of the admittance model block be denoted by the set \(\boldsymbol{\mathcal{A}_k}\), where:

\begin{align}
    \boldsymbol{\mathcal{A}_k} = \bigcup\limits_{i=1}^{\mathrm{N}} \{ \mathbf{x}^\mathrm{c}_\mathrm{a}(k+i), \theta^\mathrm{s}_\mathrm{a}(k)\}
    \label{ad_output}
\end{align}
This set is passed on as goal positions in the MPC formulation that uses I-LIP and computes desired footstep locations. 

\subsection{Walking Pattern Generation using MPC}
MPC is extensively used in the literature on legged robots for footstep planning \cite{katayama2023model}. 
 Its predictive decision-making capability, inherent constraint-handling, robustness, and ability to accommodate multi-objective optimization make it a strong candidate for this task.  

In this subsection, we formulate the walking pattern generator for physical human–humanoid collaboration as an MPC problem. The MPC is posed as a QP subject to constraints on foot placement and CoM states. Given the initial CoM states of both the object and the robot, the objective is to determine the \(N\) future optimal foot placements of the I-LIP model that enable accurate tracking of the desired robot CoM positions.  

As illustrated in Fig.~\ref{complete_frame_tro}, the MPC receives as inputs the current states \((\mathbf{X}^\mathrm{c}_k, \mathbf{X}^\mathrm{b}_k)\), the goal states \(\boldsymbol{\mathcal{A}_k}\), and the stiffness parameters \(\mathbf{K}_t\). It then outputs the \((x,y)\) position of the stance foot for the next step, \(\mathbf{u}_k\).
The formulation is as follows:\\
\vspace{0.03in}
\hrule 
\vspace{-0.06in}
\begin{align}
    &\minimize_{\mathbf{X}^\mathrm{c}_{i+1}, \mathbf{u}_i}\quad \sum_{i=k}^{k+N-1} J_i\big(\mathbf{x}^\mathrm{c}_{i},\mathbf{u}_{i-1}, \theta^\mathrm{s}_\mathrm{a}(i-1)\big)\\
    &\mathrm{subject\hspace{0.2cm} to}\hspace{0.89cm}\mathbf{X}^\mathrm{c}_{i+1}=f_\mathrm{c}(\mathbf{X}^\mathrm{c}_i,  \mathbf{x}^\mathrm{b}_i, \mathbf{v}^\mathrm{b}_\mathrm{d}, \mathbf{u}_i, \theta^\mathrm{s}_\mathrm{a}(i), \mathbf{K}_t, T)\label{ILIP_eq}\\
    &\hspace{2.5cm}\mathbf{u}_i=(u^x_i,u^y_i)\in \mathcal{U}_i\label{foot_bound_eq}
\end{align}
\vspace{-0.18in}
\hrule
\vspace{0.2cm}
The decision variables of the optimization are the robot’s CoM states, \(\mathbf{X}^\mathrm{c}_{i+1}\), and the stance foot positions, \(\mathbf{u}_{i}\), for all \(i \in \{k, k+1, \ldots, k+N-1\}\). It is important to note that the yaw angles are not included as decision variables; instead, the desired yaw values are provided directly as parameters of the optimization. This choice ensures that the MPC formulation remains convex.  
The optimization is subject to two constraints:  
(i) the step-to-step dynamics of the I-LIP model~\eqref{ILIP_eq}, and  
(ii) the foot placement constraint~\eqref{foot_bound_eq}.  
The cost \(J_i\) is defined as a weighted sum of two individual cost terms, \(\Phi_{k,1}\) and \(\Phi_{k,2}\), given by:

\subsubsection{Goal position tracking}
This objective term penalizes the deviation between the robot’s CoM position and the corresponding goal position, thereby promoting footstep patterns that minimize this error. The cost is defined as:
\begin{align}
    \Phi_{k,1} &= \big(\mathbf{x}^\mathrm{c}_{k} - \mathbf{x}^\mathrm{c}_\mathrm{a}(k)\big)^\top
    \mathbf{K}_{\theta^\mathrm{s}_\mathrm{a}(k-1)}
    \big(\mathbf{x}^\mathrm{c}_{k} - \mathbf{x}^\mathrm{c}_\mathrm{a}(k)\big), \\
    \mathbf{K}_{\theta^\mathrm{s}_\mathrm{a}(k-1)} &= 
    \mathbf{R}\big(\theta^\mathrm{s}_\mathrm{a}(k-1)\big)^\top 
    \mathbf{K}_\Phi \,
    \mathbf{R}\big(\theta^\mathrm{s}_\mathrm{a}(k-1)\big),
\end{align}
where \(\mathbf{K}_\Phi\) is a positive-definite diagonal gain matrix. The gains are first selected in the local $X'$ (forward) and $Y'$ (lateral) directions, and then rotated into the world frame using the current stance foot yaw \(\theta_\mathrm{a}^\mathrm{s}(k-1)\). This formulation enables independent tuning of stability in the forward and lateral directions.

\subsubsection{CoM-foot distance regulation}
The second cost term penalizes the distance between the robot’s CoM and the stance foot at the end of the step. This discourages footstep placements that lead to increased distance between CoM and the foot, which can cause the robot to fall. A similar principle appears in the capture point concept for the LIP~\cite{Koolen2012Capturability}, where a larger separation between the capture point and the foot corresponds to a state closer to a fall.  

For our case, the cost is defined as:
\begin{align}
    \Phi_{k,2} &= \big(\mathbf{x}^\mathrm{c}_{k} - \mathbf{u}_{k-1}\big)^\top
    \mathbf{B}_{\theta^\mathrm{s}_\mathrm{a}(k-1)}
    \big(\mathbf{x}^\mathrm{c}_{k} - \mathbf{u}_{k-1}\big), \\
    \mathbf{B}_{\theta^\mathrm{s}_\mathrm{a}(k-1)} &= 
    \mathbf{R}\big(\theta^\mathrm{s}_\mathrm{a}(k-1)\big)^\top
    \mathbf{B}_\Phi \,
    \mathbf{R}\big(\theta^\mathrm{s}_\mathrm{a}(k-1)\big),
\end{align}
where \(\mathbf{B}_\Phi\) is a positive-definite diagonal gain matrix.  

The cost \(J_i\) in the MPC objective is then given by:
\[
    J_i\big(\mathbf{x}^\mathrm{c}_i, \mathbf{u}_{i-1}\big) 
    = \phi_1 \, \Phi_{k,1} + \phi_2 \, \Phi_{k,2},
\]
where \(\phi_1\) and \(\phi_2\) are scalar weights chosen according to the relative importance of each term. Once the MPC problem is solved, the next planned footstep location \(\mathbf{u}_k\) and orientation \(\theta^\mathrm{s}_\mathrm{a}(k)\) are sent to the low-level controller for foot placement tracking, as shown in Fig.~\ref{complete_frame_tro}.

\subsection{Whole Body Control using a Quadratic Program}
The objective of the WBC, or the low-level control, is to execute the high-level plan generated by the MPC on Digit. We formulate the WBC as a QP subject to a set of constraints and objectives. The objectives encompass both task-space and joint-space goals, such as swing foot trajectory tracking, torso orientation regulation, and other tasks described in detail below. The constraints ensure that the resulting control commands are consistent with the robot’s dynamics and comply with any additional bounds imposed on the states or decision variables.  

As shown in Fig.~\ref{complete_frame_tro}, the QP receives as inputs the current state of the system \((\mathbf{q}, \mathbf{v})\), the forces applied by the human, and the footstep plan \((\mathbf{u}_k, \theta^\mathrm{s}_\mathrm{a}(k))\) provided by the MPC. The solution of the QP yields the joint torque commands \(\boldsymbol{\tau}\), which are then sent to Digit for execution.

In the subsections that follow, we first provide a description of the humanoid robot Digit and define the system's state space \((\mathbf{q}, \mathbf{v})\). We then present the constraints and objectives used in the formulation of the low-level QP.

\subsubsection{Coupled System: Digit and the Object}
\begin{figure} 
\centerline{\includegraphics[width=3.6in]{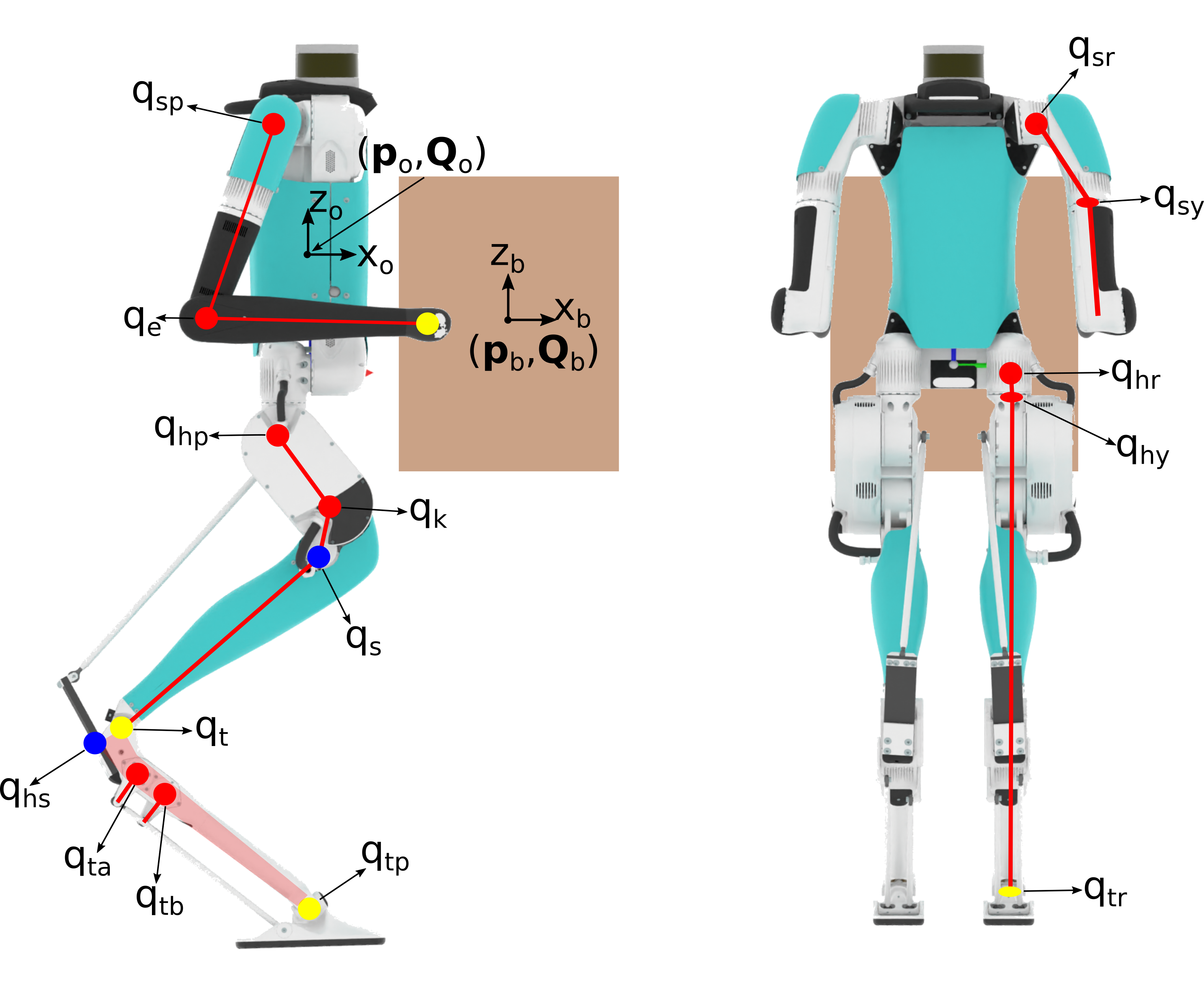}}
\caption{Configuration space of the coupled sytem comprising of Digit and the box.}
\label{digit}
\end{figure}
Digit is modeled as a tree structure composed of rigid bodies connected through joints. Its structure can be divided into five major assemblies: torso, left arm, left leg, right arm, and right leg. The overall joint configuration is illustrated in Fig.~\ref{digit}. The torso, arms, and legs have 6, 4, and 8 degrees of freedom (DoFs), respectively. In total, Digit possesses 30 DoFs, of which 20 are actuated and 4 are passive; the remaining 6 correspond to the position and orientation of the torso.  

Let the position of the torso with respect to the inertial frame be denoted by  
\(\mathbf{p}_\mathrm{o} = (x_\mathrm{o}, y_\mathrm{o}, z_\mathrm{o}) \in \mathbb{R}^3\),  
and let \(\mathbf{Q}_\mathrm{o} \in SO(3)\) represent its orientation. The torso serves as the parent body for both the legs and arms.  

Each leg, $i\in\{L,R\}$, is connected to the torso through a 3-DoF hip joint parameterized by yaw, roll, and pitch angles: \(q_\mathrm{hy}^i\), \(q_\mathrm{hr}^i\), and \(q_\mathrm{hp}^i\), respectively. This joint connects the torso to the thigh. At the distal end of the thigh, a knee joint (\(q_\mathrm{k}^i\)) and a shin spring (\(q_\mathrm{s}^i\)) connect the thigh to the shin link. The shin and the tarsus are linked via a joint parameterized by \(q_\mathrm{t}\). The heel-spring ($q_\mathrm{hs}^i$) attached to the tarsus connects to the thigh to form a compliant kinematic loop. The leg ends with a heel with two degrees of freedom described by corresponding roll ($q^i_\mathrm{tr}$) and pitch ($q^i_\mathrm{tp}$) angles. It is actuated using two four-bar mechanisms parametrized by $q_\mathrm{ta}$ and $q_\mathrm{tb}$. Finally, each arm, $i\in\{L,R\}$, attached to the torso is parametrized by shoulder yaw ($q^i_\mathrm{sr}$), roll ($q^i_\mathrm{sr}$), pitch ($q^i_\mathrm{sp}$) and elbow ($q^i_\mathrm{e}$) angles. Each leg/arm pair configuration is denoted by $\mathbf{q}_i$ as shown below.
\begin{align}
\nonumber
    \mathbf{q}_{i\in\{L,R\}}=~&(q_\mathrm{hy}^i, q_\mathrm{hr}^i, q_\mathrm{hp}^i, q_\mathrm{k}^i, q_\mathrm{s}^i, q_\mathrm{t}^i, q_\mathrm{hs}^i, q_\mathrm{tr}^i, q_\mathrm{tp}^i,\\& q_\mathrm{ta}^i, q_\mathrm{tb}^i, q_\mathrm{sy}^i, q_\mathrm{sr}^i, q_\mathrm{sp}^i, q_\mathrm{e}^i)
\end{align}
The state space of the entire free-floating Digit denoted by $(\mathbf{q}', \mathbf{v}')$ is given by:
\begin{align}
    &\mathbf{q}' = (\mathbf{p}_\mathrm{o}, \mathbf{Q}_\mathrm{o}, \mathbf{q}_\mathrm{L}, \mathbf{q}_\mathrm{R})\\&
    \mathbf{v}' = (\mathbf{\dot{p}}_\mathrm{o}, \boldsymbol{\omega}_\mathrm{o}, \mathbf{\dot{q}}_\mathrm{L}, \mathbf{\dot{q}}_\mathrm{R}) 
\end{align}
where $\boldsymbol{\omega}_o$ is the angular velocity of the torso in body frame. Let the \(\mathbf{Q}_\mathrm{b} \in SO(3)\) represent the orientation of the object. The complete state space of Digit with the states of the object is given by
\begin{align}
    &\mathbf{q} = (\mathbf{q}', \mathbf{x}^\mathrm{b}, \mathbf{Q}_\mathrm{b}) \in \mathbb{R}^{44}\\
    &\mathbf{v} = (\mathbf{v}', \mathbf{\dot{x}}^\mathrm{b}, \boldsymbol{\omega}_\mathrm{b}) \in \mathbb{R}^{42}
\end{align}
where $\boldsymbol{\omega}_b$ is the angular velocity of the box in the inertial frame.
\subsubsection{Constraints}
The literature on whole-body control for pHHI typically treats interaction forces as external disturbances within the system dynamics, without explicitly planning for these forces. In contrast, our approach models the coupled dynamics of the robot (Digit) and the object, enabling the robot to actively regulate the interaction forces. This formulation also provides the flexibility to control both the position and orientation of the object.  

In our walking model, locomotion alternates between left and right single-support phases, with the double-support phase assumed to have negligible duration. The dynamics of Digit during the single-support phase are expressed as:
\begin{align}
    \mathbf{D}(\mathbf{q})\mathbf{\dot{v}} + \mathbf{C}(\mathbf{q},\mathbf{v}) =& \mathbf{S}^\top_\tau \boldsymbol{\tau} + \big(\mathbf{J}(\mathbf{q})\big)^\top\boldsymbol{\lambda} \label{Digit_dyn_const} \\\nonumber&+ \big(\mathbf{J}_\mathrm{L}(\mathbf{q})\big)^\top\boldsymbol{\lambda}_\mathrm{L}+\big(\mathbf{J}_\mathrm{R}(\mathbf{q})\big)^\top\boldsymbol{\lambda}_\mathrm{R}
\end{align}
where \(\mathbf{D}(\mathbf{q})\) is the inertia matrix, \(\mathbf{C}(\mathbf{q},\mathbf{v})\) is the matrix containing velocity-dependent terms and gravitational forces, and \(\mathbf{S}_\tau\) is the selection matrix for the actuated joints. \(\mathbf{J}_\mathrm{L}(\mathbf{q})\) denotes the Jacobian associated with the ball-joint constraint between the left hand and the object, and \(\boldsymbol{\lambda}_\mathrm{L} = [\lambda^\mathrm{x}_\mathrm{L}, \lambda^\mathrm{y}_\mathrm{L}, \lambda^\mathrm{z}_\mathrm{L}]^\top \in \mathbb{R}^3\) is the corresponding generalized force. An analogous notation is used for the right hand, replacing the subscript `L' with `R'.  
\(\mathbf{J}(\mathbf{q})\) denotes the Jacobian of all holonomic constraints not involving the object, and \(\boldsymbol{\lambda} \in \mathbb{R}^{16}\) are the associated generalized forces. These constraints include:  
(a) six closed-loop kinematic constraints arising from the Digit’s mechanical structure,  
(b) a stance-foot contact constraint, and  
(c) stiff spring constraints.  
The closed-loop kinematic constraints are handled in a manner similar to the approach used for the Cassie robot~\cite{Ames2021UInverse}.
All these constraints are incorporated as:
\begin{align}
    \mathbf{J}(\mathbf{q})\mathbf{\dot{v}} &+ \mathbf{\dot{J}}(\mathbf{q},\mathbf{v})\mathbf{v} = \mathbf{0}
    \label{robot_const}
\end{align}
Next, we incorporate the dynamics of the object, which is modeled as a single rigid body. Let \(\boldsymbol{\alpha}_\mathrm{b}\) and \(\boldsymbol{\omega}_\mathrm{b}\) denote the angular acceleration and angular velocity of the object, respectively, expressed in the world frame. The object’s moment of inertia matrix in the body frame is denoted by \(\mathbf{I}_\mathrm{b}\), and its mass by \(m_\mathrm{b}\).  
Let \(\mathbf{F}_\mathrm{h}\) and \(\mathbf{M}_\mathrm{h}\) represent the net forces and moments applied by the human at the object’s CoM in the world frame. The rotation matrix \(\mathbf{R} \in SO(3)\) transforms quantities from the object frame to the world frame, and \(\mathbf{F}_\mathrm{g}\) denotes the gravitational force.  
Finally, let \(\mathbf{r}_\mathrm{L}\) and \(\mathbf{r}_\mathrm{R}\) be the vectors, expressed in the world frame, from the object’s CoM to the left-hand and right-hand contact points, respectively. The dynamics of the object can then be expressed as:
\begin{align}
    \mathbf{M}_\mathrm{h} -\mathbf{r}_\mathrm{L}\times\boldsymbol{\lambda}_\mathrm{L}-\mathbf{r}_\mathrm{R}\times\boldsymbol{\lambda}_\mathrm{R} =&~ \mathbf{R}\mathbf{I}_\mathrm{b}\mathbf{R}^T\boldsymbol{\alpha}_\mathrm{b} \nonumber\\&+ \boldsymbol{\omega}_\mathrm{b}\times \mathbf{RI}_\mathrm{b}\mathbf{R}^T \boldsymbol{\omega}_\mathrm{b}\label{box_rot_dyn_const}
\end{align}
\vspace{-0.32in}
\begin{align}
    m_\mathrm{b}\mathbf{\ddot{x}}^\mathrm{b} = \mathbf{F}_\mathrm{g} -\boldsymbol{\lambda}_\mathrm{L}-\boldsymbol{\lambda}_\mathrm{R} +\mathbf{F}_\mathrm{h}
    \label{box_xyz_dyn_const}
\end{align}
Finally, we add 2 ball joint constraints between the two arms and the object:
\begin{align}
    \mathbf{J}_i(\mathbf{q})\mathbf{\dot{v}}+\mathbf{\dot{J}}_i(\mathbf{q}, \mathbf{v})\mathbf{v} =& \mathbf{\ddot{x}}^\mathrm{b}+ \mathbf{r}_i\times\boldsymbol{\alpha}_\mathrm{b}\label{arm_box_const}\\\nonumber& + \boldsymbol{\omega}_\mathrm{b}\times(\boldsymbol{\omega}_\mathrm{b}\times\mathbf{r}_i)~\forall~i\in\{\mathrm{L},\mathrm{R}\}
\end{align}
This completes the constraints regarding modeling the coupled system.
Next, we require the foot contact wrench, $(\lambda^x, \lambda^y, \lambda^z, \lambda^{mx}, \lambda^{my}, \lambda^{mz})$ to lie inside the linearized Contact Wrench Cone (CWC) $\mathcal{K}$ \cite{Caron2015Stability}. We also bound the input torque and interaction forces at the hands as:
\begin{align}
    \mathbf{lb}_{\tau}&\leq \boldsymbol{\tau} \leq \mathbf{ub}_{\tau}\label{tau_bound}\\
    \mathbf{lb}_\mathrm{h} & \leq \boldsymbol{\lambda}_i \leq \mathbf{ub}_\mathrm{h}~\forall ~i\in\{L,R\}\label{force_bound} 
\end{align}
Here, $\mathbf{lb}_{\tau}$ and $\mathbf{ub}_{\tau}$ are the lower and upper bounds on the input torque, $\boldsymbol{\tau}$. Moreover, $\mathbf{lb}_\mathrm{h}$ and $\mathbf{ub}_\mathrm{h}$ are the lower and upper bounds on the interaction forces between the box and the hand.
\subsubsection{Trajectory Tracking Objectives}
We now describe the objectives related to tracking the desired trajectories of various quantities of the Digit humanoid. These quantities include the trajectory of the swing foot, the orientation of the swing foot, the orientation of the torso, and the partial orientation of the object. Let the errors between the desired and actual values of these quantities be stacked into the vector \(\mathbf{e} \in \mathbb{R}^{11}\). The corresponding tracking objective, commonly adopted in the literature~\cite{Poulakakis2022Sequential}, is to minimize the cost function \(\Psi_1\), defined as:
\begin{align}
    \Psi_1 = ||\mathbf{\ddot{e}}+\mathbf{K}_\mathrm{D}\mathbf{\dot{e}}+\mathbf{K}_\mathrm{P}\mathbf{e}||^2
\end{align}
where, $\mathbf{K}_\mathrm{D}$ and $\mathbf{K}_\mathrm{P}$ are appropriately selected gain matrices. 

To align the torso orientation with the estimated intended yaw of the object, we define the desired torso orientation as  
\[
\mathbf{Q}^\mathrm{torso}_\mathrm{des} = Q(0, 0, \theta^\mathrm{b}_\mathrm{d}),
\]  
where \(Q(\cdot)\) denotes the function that converts roll–pitch–yaw angles into a quaternion representation.

Using the MPC plan for the next step ($\mathbf{u}_k, \theta^\mathrm{s}_\mathrm{a}(k)$), we get the desired swing foot trajectory as:
\begin{align}
    \mathbf{p}_\mathrm{sw}^\mathrm{des}(s) = \begin{bmatrix}
        p_\mathrm{sw}^\mathrm{x}(k-1)+\frac{1}{2}(1-\cos\pi s)u^\mathrm{x}_\mathrm{k}\\
        p_\mathrm{sw}^\mathrm{y}(k-1)+\frac{1}{2}(1-\cos\pi s)u^\mathrm{y}_\mathrm{k}\\
        P(s,z_\mathrm{cl})
    \end{bmatrix}
\end{align}
Here, $s$ is a phase variable linearly varying from 0 at the start of the step to 1 at the end of the step.
$P$ is a fourth-order polynomial in $s$, designed to ensure that the swing foot starts with zero initial vertical velocity and zero initial vertical position with respect to the stance foot, while attaining a maximum height of $z_\mathrm{cl}$ during the step. $  \mathbf{p}_\mathrm{sw}(k) = \begin{bmatrix}
    p_\mathrm{sw}^\mathrm{x}(k) & p_\mathrm{sw}^\mathrm{y}(k) & p_\mathrm{sw}^\mathrm{z}(k)
\end{bmatrix}^T$ is the swing foot position at the start of the $k^\mathrm{th}$ step. For the orientation of the swing foot, we set the desired orientation $\mathbf{Q}^\mathrm{sw}_\mathrm{des} = 
Q\big(0,0,\theta_\mathrm{a}^\mathrm{s}(k)\big)$,
where \(Q(\cdot)\) converts roll–pitch–yaw angles into quaternion form.

The robot is selectively commanded to control the roll and pitch of the object, maintaining both at a desired value of zero. All orientation errors are expressed in the global frame. The desired quaternion for the object is defined as \(Q(0,0,0)\); however, the weight associated with the \(z\)-axis rotation is set to zero. 
This formulation implicitly governs the object’s vertical position based on the human’s input: when the human raises or lowers the box, a pitch error is detected, prompting the robot to adjust the height accordingly. Rotation about the $z$-axis is handled separately through a dedicated objective that enforces compliance in yaw.

\subsubsection{Other Objectives}
We impose objectives on both the humanoid and object center-of-mass (CoM) accelerations to achieve the compliant behavior prescribed by the admittance model. This formulation allows the robot’s arms to actively contribute to compliance, complementing the stepping pattern generated by the MPC. Let \(\mathbf{a}^\mathrm{c}\) and \(\mathbf{a}^\mathrm{b}\) denote the desired CoM accelerations of the humanoid and the object, respectively. These quantities are obtained from the admittance model depicted in Fig.~\ref{all_figures}(b) and are computed as:
\begin{align}
    \nonumber\mathrm{m}_\mathrm{c}\mathbf{a}^\mathrm{c} &= \mathbf{R}\left(\theta^\mathrm{s}\right)\mathbf{K}_\mathrm{h}\big(\mathbf{R}\left(\theta^\mathrm{s}\right)\big)^\top\big(\mathbf{x}^\mathrm{b}-\mathbf{x}^c-\mathbf{x}^d\big)\\&+ \mathbf{R}(\theta^\mathrm{s})\mathbf{B}_\mathrm{h}\big(\mathbf{R}(\theta^\mathrm{s})\big)^\top\big(\mathbf{\dot{x}}^\mathrm{b}-\mathbf{\dot{x}}^\mathrm{c}\big)\\\nonumber
    \mathrm{m_b}\mathbf{a}^\mathrm{b} &= \mathbf{F}_\mathrm{h}-\mathbf{R}(\theta^\mathrm{s})\mathbf{K}_\mathrm{h}\big(\mathbf{R}(\theta^\mathrm{s})\big)^\top\big(\mathbf{x}^\mathrm{b}-\mathbf{x}^\mathrm{c}-\mathbf{x}^\mathrm{d}\big) \\&- \mathbf{R}(\theta^\mathrm{s})\mathbf{B}_\mathrm{h}\big(\mathbf{R}(\theta^\mathrm{s})\big)^\top\big(\mathbf{\dot{x}}^\mathrm{b}-\mathbf{\dot{x}}^\mathrm{c}\big)
\end{align}
Here, $\mathbf{K}_\mathrm{h} = diag(K^x_\mathrm{h}, K^y_\mathrm{h}) $ and $\mathbf{B}_\mathrm{h} = diag(B^x_\mathrm{h}, B^y_\mathrm{h})$ are different from 
those in the admittance model,
as we hypothesize that low-compliance desired trajectories at the high level and high-compliance desired trajectories at the low level lead to more efficient collaboration. We prove this hypothesis in the results section of this paper. The apparent inertia experienced by the human can be modified by adjusting the masses of the object and the robot in these equations. The tracking objective for the humanoid CoM acceleration is defined as  
\begin{equation}
    \Psi_2 = \left\| \mathbf{\ddot{x}}^\mathrm{c} - \mathbf{a}^\mathrm{c} \right\|^2 ,
\end{equation}
while the corresponding objective for the object CoM acceleration is given by  
\begin{equation}
    \Psi_3 = \left\| \mathbf{\ddot{x}}^\mathrm{b} - \mathbf{a}^\mathrm{b} \right\|^2 .
\end{equation}
In most frameworks, the desired CoM acceleration for the low-level controller is derived from the simplified model used for planning \cite{Kumbhar2025MPC}. In this work, however, we instead use the accelerations from the admittance model as the desired CoM accelerations, thereby promoting \emph{local} compliance. This choice is motivated by the fact that the high-level planner enforces a desired compliance objective, whereas the stiffness modulation only attempts to track the desired distance between the humanoid and the object. Consequently, the desired accelerations generated by the I-LIP model do not exclusively promote compliance. We have similar costs on the yaw acceleration of the object given by:
\begin{equation}
    \Psi_4 = \left\| \mathbf{\ddot{\theta}}^\mathrm{b} - \alpha^\mathrm{b} \right\|^2 .
\end{equation}
Here, $\alpha^\mathrm{b}$ is the desired acceleration given by the admittance model as:
\begin{align}
    \alpha^\mathrm{b} = k^{\mathrm{b}}_\mathrm{P}(\theta^\mathrm{s}-\theta^\mathrm{b}) - k^\mathrm{b}_\mathrm{D}\dot{\theta}^\mathrm{b}
\end{align}
Here, $k^{\mathrm{b}}_\mathrm{P}$ and $k^\mathrm{b}_\mathrm{D}$ are positive constants for modulating the stiffness in the rotational $z$ direction. We add the small damping term $k^\mathrm{b}_\mathrm{D}\dot{\theta}^\mathrm{b}$ to stabilize the response.
The CoM in the I-LIP model is represented as a point mass and, therefore, cannot capture the effects of angular momentum on Digit. We, therefore, minimize the angular momentum as done in \cite{Poulakakis2022Sequential}. We compute the angular momentum of the digit about its CoM, $\mathbf{\eta}_2 = \mathbf{A}_\mathrm{ang}\mathbf{v}$. Here, $\mathbf{A}_\mathrm{ang}$ is the part of the centroidal momentum matrix. The minimization objective is given by:
\begin{align}
    \Psi_5 = ||\mathbf{\dot{\eta}}_2+ \mathbf{K}_\mathrm{ang}\mathbf{\eta}_2||^2 .
\end{align}
$\mathbf{K}_\mathrm{ang}$ is appropriately selected gain matrix.
Next, we also minimize the input torque as:
\begin{align}
    \Psi_6 = ||\mathbf{\tau}||^2
\end{align}
Finally, we minimize the x and y direction forces at the point of interaction between the box and the arms of the robot.
\begin{align}
    \Psi_7 = ||\mathbf{\lambda}^\mathrm{L}_\mathrm{xy}||^2+ ||\lambda^\mathrm{R}_\mathrm{xy}||^2
\end{align}
\subsubsection{Weighted QP}
To formulate the optimization problem, we construct a single quadratic cost function by taking a weighted sum of the objectives described above. Together with the system constraints, this cost function defines the QP, expressed as:
\\
\vspace{-0.1in}
\hrule
\vspace{-0.06in}
\begin{align}    &\minimize_{\mathbf{\dot{v}},\mathbf{\ddot{x}}_b, \boldsymbol{\alpha}_\mathrm{b}, \boldsymbol{\tau}, \boldsymbol{\lambda}, \boldsymbol{\lambda}^\mathrm{L},\boldsymbol{\lambda^\mathrm{R}}}\quad  \sum_{i=1}^{7} \psi_i\Psi_i\\
    &\mathrm{subject\hspace{0.2cm} to}\quad\quad \eqref{Digit_dyn_const}, \eqref{robot_const}, \eqref{box_rot_dyn_const}, \eqref{box_xyz_dyn_const}, \eqref{arm_box_const}, \eqref{tau_bound}, \eqref{force_bound}\nonumber\\
    &\hspace{2.4cm}(\lambda^x,\lambda^y,\lambda^z,\lambda^{mx},\lambda^{my},\lambda^{mz})\in \mathcal{K}\
\end{align}
\vspace{-0.198in}
\hrule
\vspace{0.2cm}
The weighting gains \(\psi_i,~\forall i \in \{1,2,3,4,5\}\) are chosen based on the relative priority of the corresponding objectives. The final constraint enforces that the contact wrench at the stance foot lies within the linearized contact wrench cone \(\mathcal{K}\).
\subsection{Stiffness Modulation}
The springs and dampers in the I-LIP model apply forces to the robot’s CoM during the footstep planning phase, thereby adjusting the planned footstep pattern based on these interaction forces. We hypothesize that using constant spring and damper parameters will not yield the desired compliant behavior between the object and the robot, due to discrepancies between the forces actually experienced by the CoM and those predicted by the I-LIP model. Authors in \cite{Kumbhar2025MPC} prove this hypothesis for straight line walking. However, their approach employs a global trajectory from the admittance model for stiffness modulation, which does not produce \emph{local} compliance. In this work, we modify the stiffness modulation to achieve convergence to a desired distance between the humanoid and the object rather than to a global trajectory. While this does not fully realize local compliance, the modulation now operates toward convergence to the desired distance ($x^\mathrm{d}_x$) prescribed by the local compliance objective. Additionally, we adopt this strategy for considering turning maneuvers as well.

The modulation of the stiffness of the \(x\)-component spring at each iteration of the low-level control loop is given by:
\begin{align}
    K^\mathrm{x}_t &= K^\mathrm{x}_t - k^\mathrm{x}_1\big(x^\mathrm{b}_l(t) - x^\mathrm{c}_l(t) - x^\mathrm{d}\big)\\\nonumber& - b^\mathrm{x}_1\big(\dot{x}^\mathrm{b}_l(t)-\dot{x}^\mathrm{c}_l(t)\big)  \\
    \mathbf{x}^i_l = &\big(\mathbf{R}(\theta^s)\big)^\top\mathbf{x}^i  \quad \dot{\mathbf{x}}^i_l =  \big(\mathbf{R}(\theta^s)\big)^\top\dot{\mathbf{x}}^i \quad i\in\{\mathrm{b},\mathrm{c}\}
\end{align}
The objective of this modulation is not to precisely estimate the forces experienced by the CoM, but rather to adjust them such that the stepping pattern generated by the MPC achieves the desired compliant behavior while promoting convergence of the distance between the CoMs of the robot and the object to the target value \(x^\mathrm{d}_\mathrm{x}\). This approach is analogous to concepts in adaptive control theory, where the primary goal is to accomplish the task while estimating unknown parameters, with the understanding that these estimates may differ from their true values.

\section{RESULTS}
We validate our proposed framework through both simulation studies and real-world experiments\footnote{A video of the experimental demonstrations is available at: \\https://www.youtube.com/watch?v=VeCh9ycYWWA} on the humanoid platform Digit. The evaluation spans multiple collaborative locomotion and manipulation tasks, including in-place walking under varying compliance levels, two-dimensional translations, and turning maneuvers. Combinations of these basic maneuvers can be used to realize more complex behaviors such as navigation in cluttered environments or operation within structured workspaces. Through our framework, these maneuvers are autonomously executed without needing thresholding or state machines.

The control framework is first tested in the MuJoCo physics engine \cite{Todorov2012Mujoco} and then deployed on the real robot. To reduce the sim-to-real gap, we add actuator dynamics through the MuJoCo plugin, include joint friction and damping, inject observation noise, introduce delays in control commands, and vary spring parameters in the humanoid model.

Together, these measures significantly improved the fidelity of the simulation environment. As a result, controllers designed and validated in MuJoCo required only minimal tuning when transferred to the physical robot. All results presented in this work are therefore based on successful execution on the real Digit platform, with simulation serving as a critical intermediary step for safe, systematic, and resource-efficient validation.

\begin{figure}
    \centering
    \includegraphics[width=0.9\linewidth]{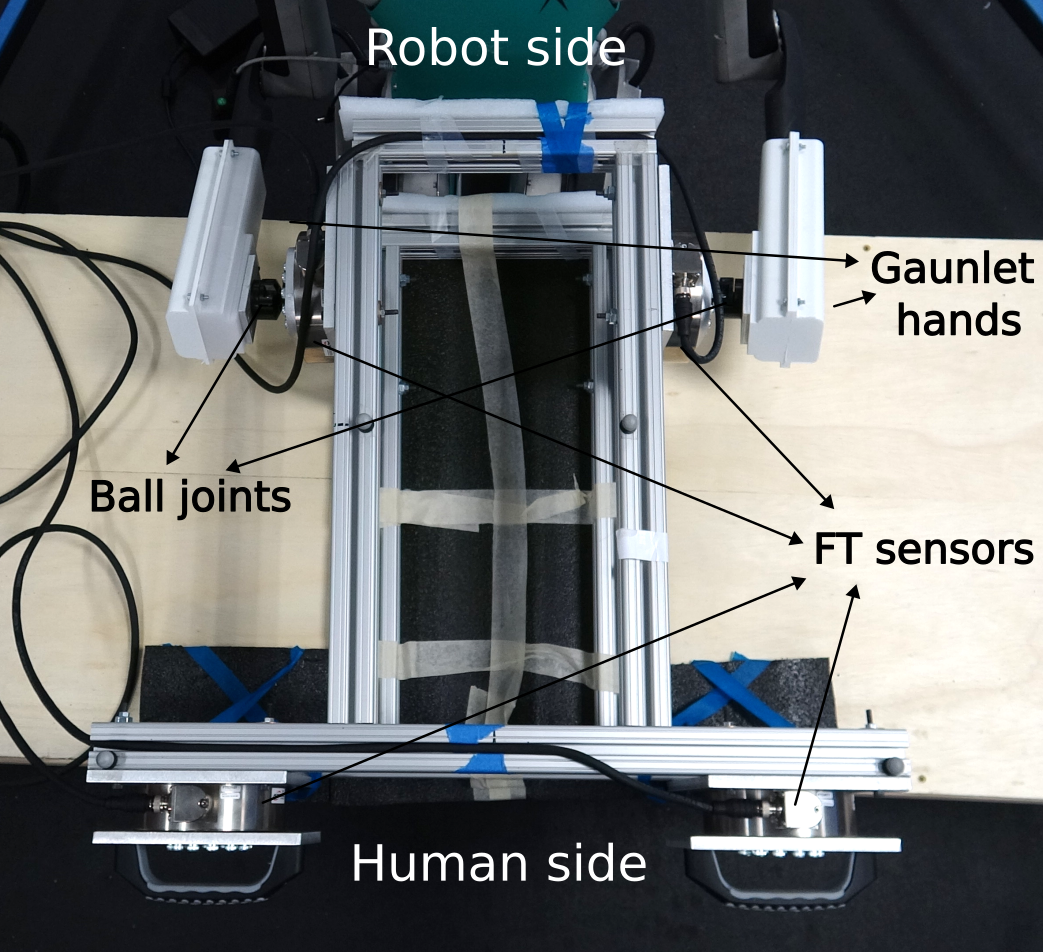}
    \caption{Physical box used for the collaboration task. The box itself is made of aluminum and weighs around 15 kg. The hands of Digit are connected to the box through a 3D-printed gauntlet and stainless steel ball joint assembly. The interaction ports at the human and robot side also consist of force torque sensors.}
    \label{box}
\end{figure}

This version of Digit lacks wrists to emulate ball joints. To provide wrist functionality and enable interaction force measurement, we designed an aluminum box, housing four ATI Axia130 M125 force–torque sensors: two on the human side, terminating in handles, and two on Digit's side, connected through a ball-joint and gauntlet design, as shown in Fig. \ref{box}.
The box weighs 15 kg or 33 pounds. Additionally, we use a camera-based motion tracking system and related software (Optitrack Motive 3.1.1) for pose and velocity observations of the box using reflective markers. For the humanoid, we obtain the linear velocity of the torso from the motion capture system, while the rest of the states come from the in-built estimator. 

In the following subsections, we present the analysis and main contributions of our framework in the context of co-transportation tasks involving heavy objects.

\subsection{In-place Walking and Joint Manipulation Evaluation}
\begin{figure}
    \centering
    \includegraphics[width=0.98\linewidth]{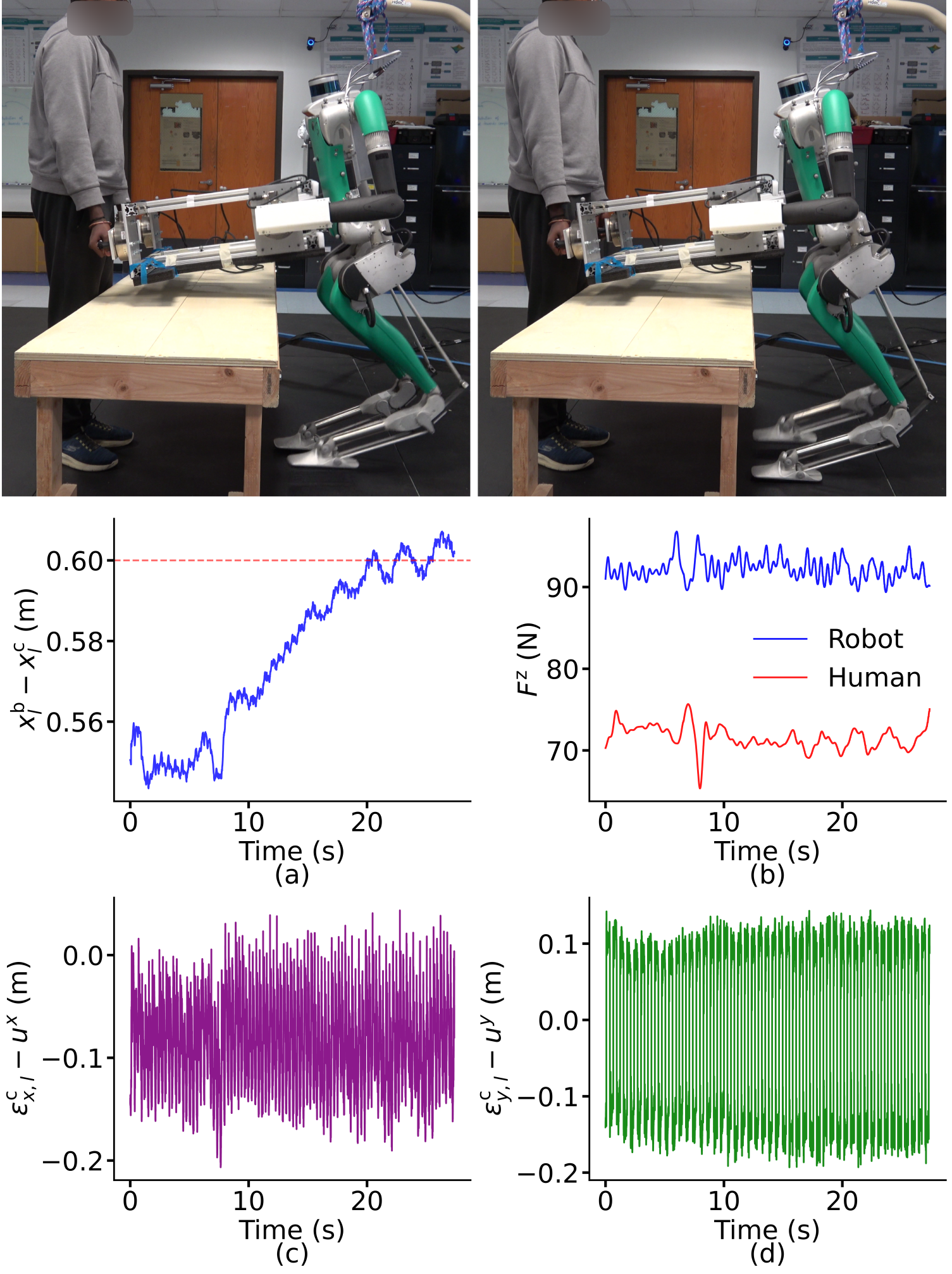}
    \caption{Results on collaborative in-place walking task. Top: snapshots show the humanoid sharing the load while stepping in place. Bottom: the subscript \emph{l} in the variables indicates representations in the local frame about the stance foot ($X'$, $Y'$). (a) The distance between the CoMs of the humanoid and the object ($x^\mathrm{b}_l - x^\mathrm{c}_l$) converges to the desired value of 0.6 m. (b) Vertical direction forces applied by the human and the robot ($F^z_\mathrm{h}, F^z_\mathrm{r}$) on the object. (c,d) Modified capture point offsets in the locally forward ($\epsilon^\mathrm{c}_{x,l}$) and lateral direction ($\epsilon^\mathrm{c}_{y,l}$) remain bounded, indicating stability. }
    \label{inplace}
\end{figure}
\begin{figure*}
    \centering
    \includegraphics[width=0.9\linewidth]{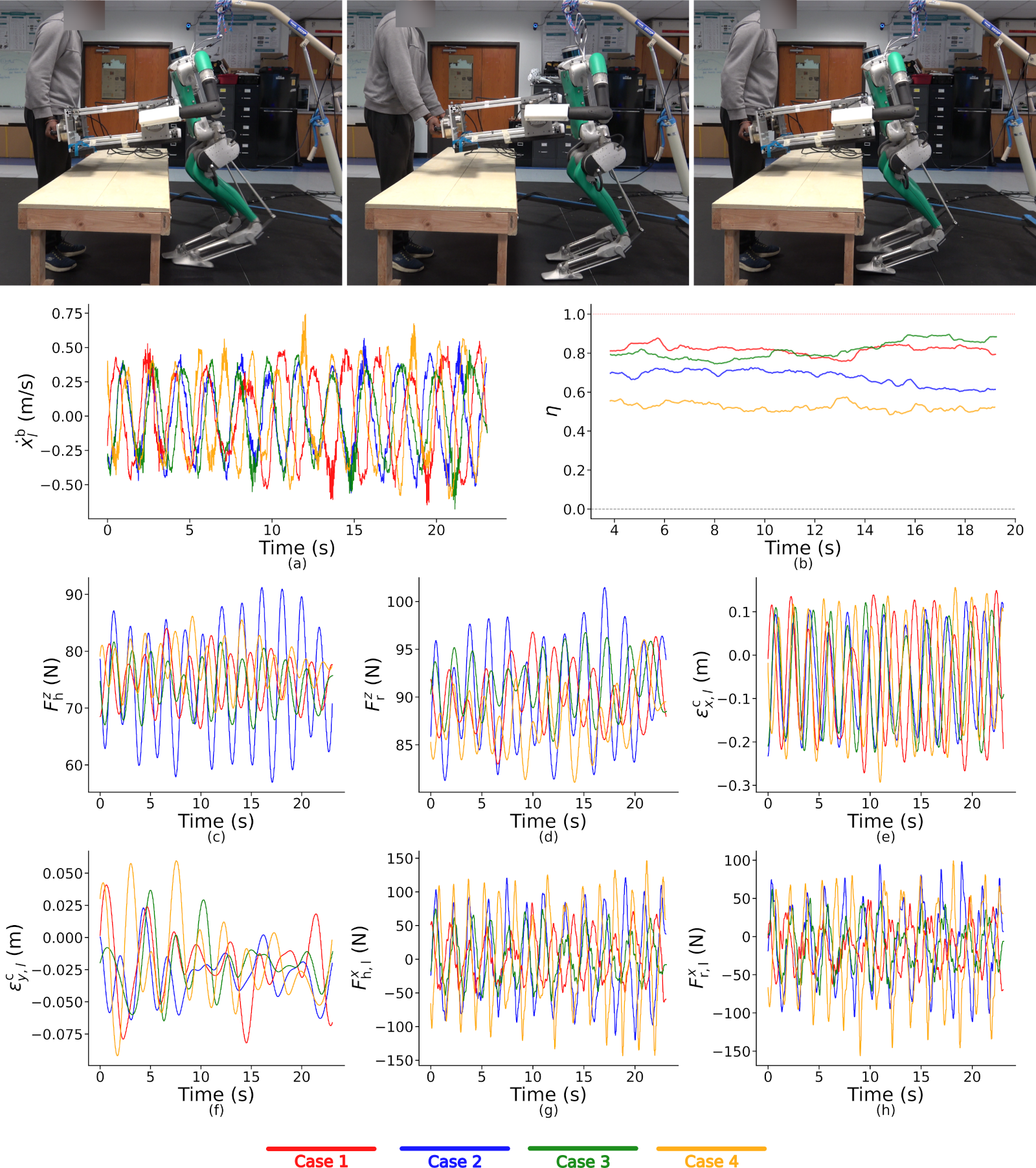}
    \caption{Top: Experiment illustrating the periodic motion task in which the human moves the box forward and backward in a periodic fashion, as shown in the task progression snapshots. Bottom: The subscript \emph{l} indicates representations in the local frame about the stance foot ($X'$, $Y'$). (a) Box velocity ($\dot{x}^\mathrm{b}_l$) in the forward direction for all four cases showing the periodic motion. (b) Efficiency ($\eta$) of all four cases, showing high efficiencies in cases involving low-compliance objectives for the stepping pattern. (c,d) Human and humanoid's vertical force ($F^z_\mathrm{h}$ and $F^z_\mathrm{r}$) on the box, showing more than 50\% contribution from the robot. (e,f) Modified capture point offsets in the forward direction ($\epsilon^\mathrm{c}_{x,l}$) and the lateral direction ($\epsilon^\mathrm{c}_{y,l}$). It remains bounded, indicating that the stepping pattern is able to capture the robot and prevent falls. (g,h) Forces applied by the human ($F^x_\mathrm{h}$) and the humanoid ($F^x_\mathrm{r}$) in the forward direction.}
    \label{periodic}
\end{figure*}
\begin{figure}
    \centering
    \includegraphics[width=0.95\linewidth]{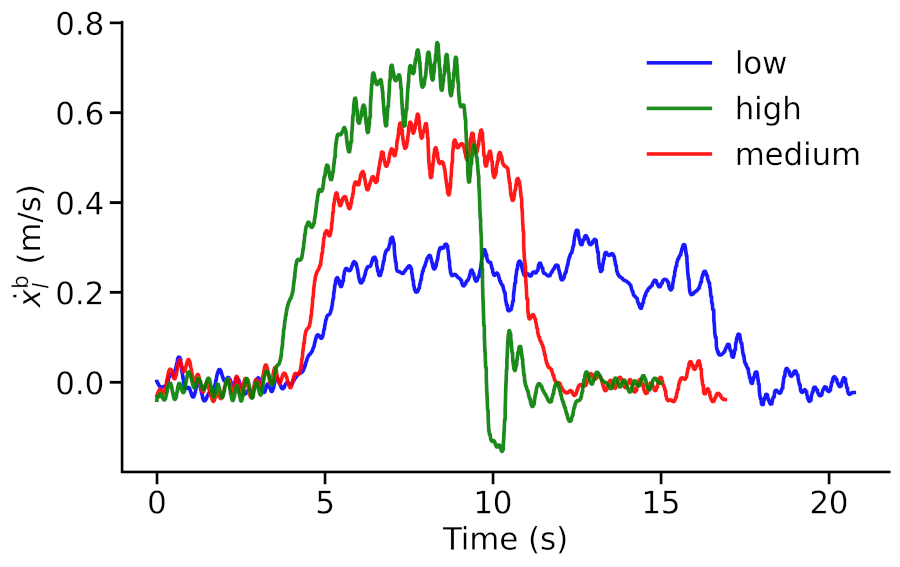}
    \caption{Different speeds of the box in the forward direction ($\dot{x}^\mathrm{b}_l$), low, medium and high, during the straight line walking task.
    }
    \label{speeds}
\end{figure}

\begin{figure*}
    \centering
    \includegraphics[width=0.85\linewidth]{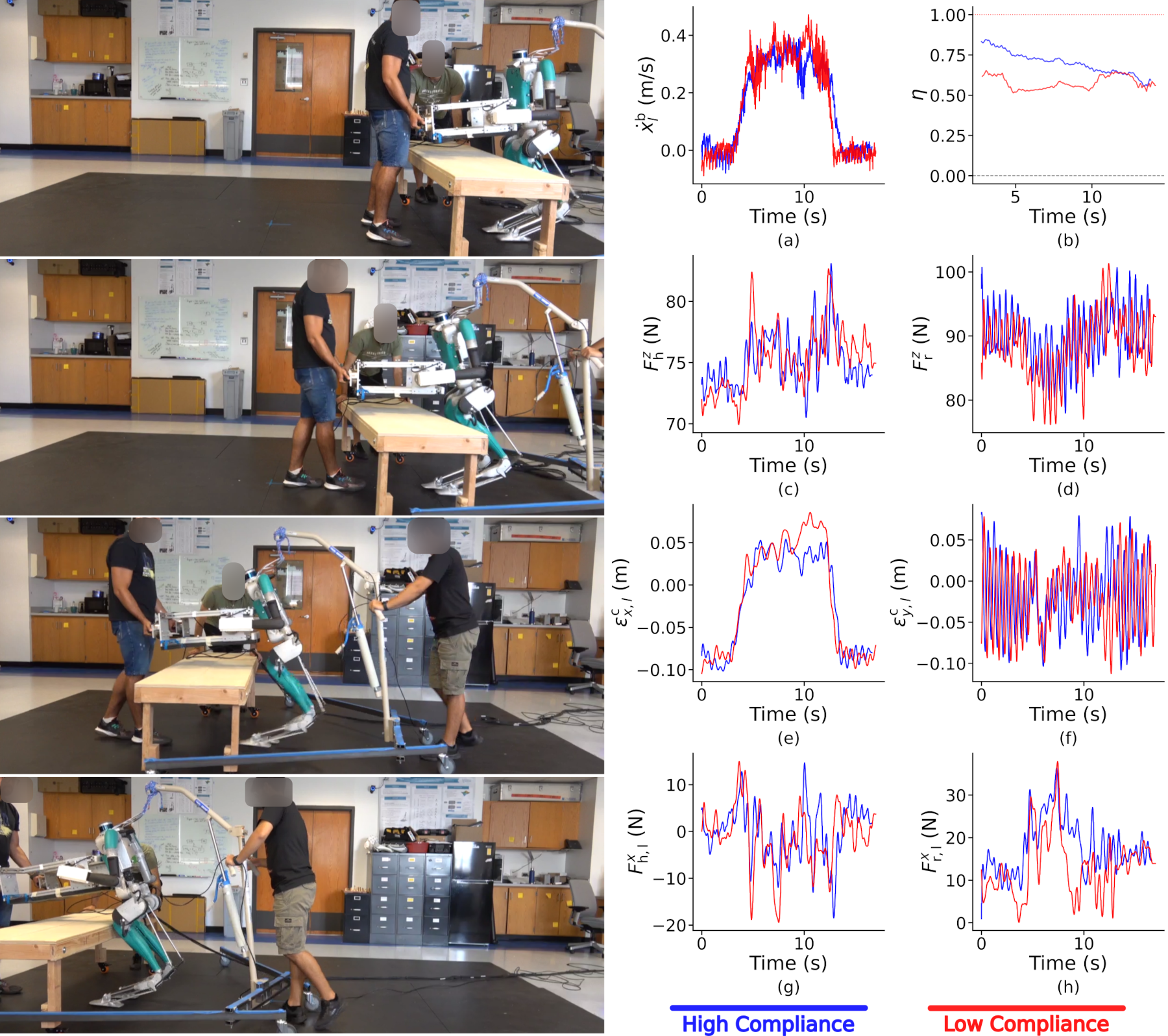}
    \caption{Experiment illustrating the straight walking task. Left: The human starts moving the box with an intended speed, and the humanoid follows it as shown in the progression through images. Right: We analyze two cases, high and low compliance in hands, for around the same velocity task. The subscript \emph{l} indicates representations in the local frame about the stance foot ($X'$, $Y'$). (a) Box velocity ($\dot{x}^\mathrm{b}_l$) in the forward direction for both cases. (b) Efficiency ($\eta$) of the task shows consistently higher values for high compliance. (c,d) Vertical forces applied by the human and the humanoid on the object ($F^z_\mathrm{h}$ and $F^z_\mathrm{r}$), showing the load sharing. (e,f) Modified capture point offsets in the forward direction ($\epsilon^\mathrm{c}_{x,l}$) and the lateral direction ($\epsilon^\mathrm{c}_{y,l}$). (g,h) Forces exerted by the human and the humanoid on the box in the forward direction ($F^x_\mathrm{h,l}$ and $F^x_\mathrm{r,l}$), which remain minimal.
    }
    \label{straight}
\end{figure*}

\begin{figure*}
    \centering
    \includegraphics[width=0.85\linewidth]{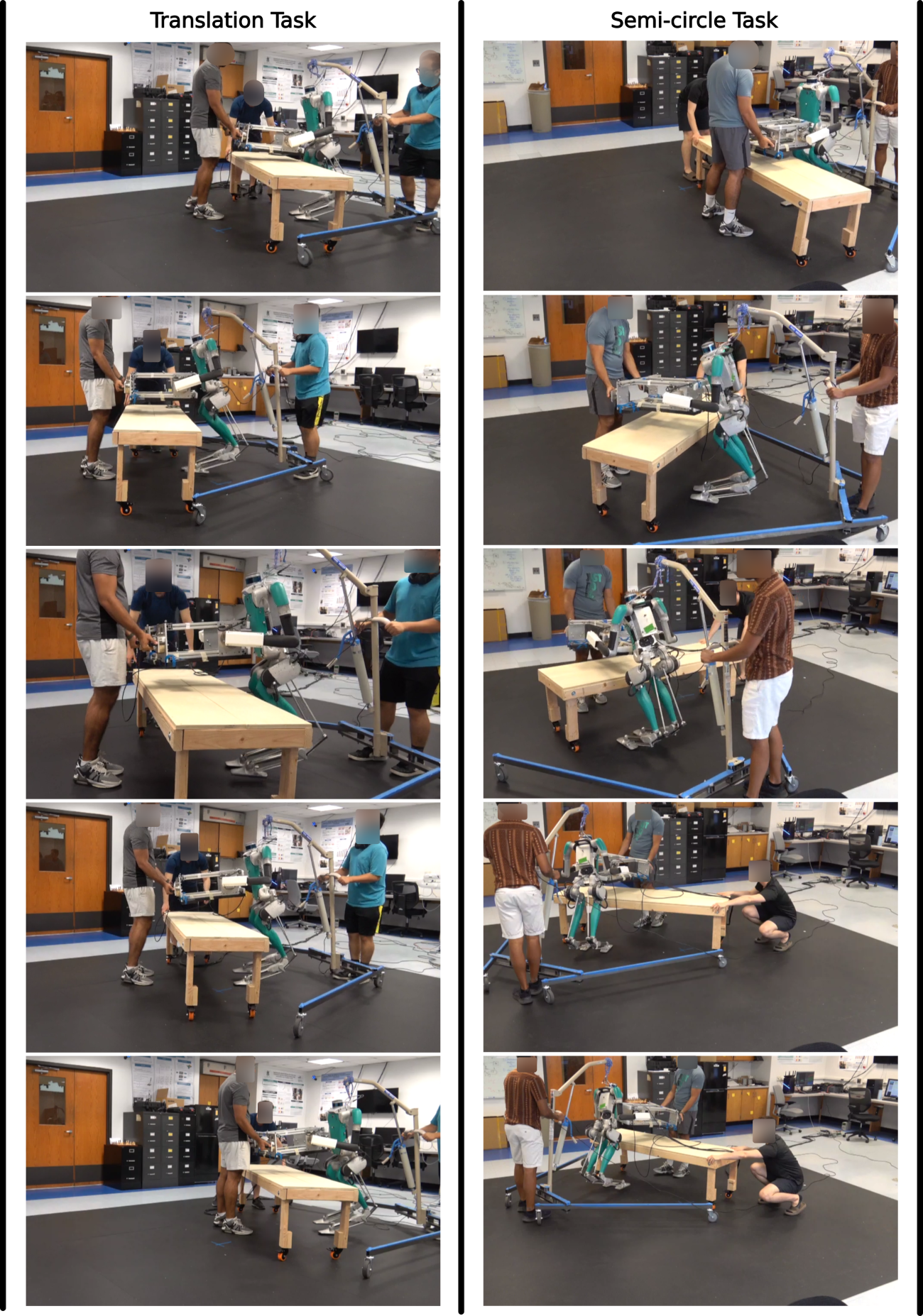}
    \caption{Progressions of the translational (left) and semi-circle task (right) over time. The humanoid is able to follow complex trajectories for co-transportation tasks.}
    \label{complex}
\end{figure*}
This task involves a human and the humanoid lifting an object and keeping it in one place. The humanoid is doing in-place walking throughout the experiment. 
Figure \ref{inplace} illustrates the progression of the experiment over time through images, with the accompanying plots presenting the analysis and results. Through these experiments, we aim to demonstrate the following aspects of the framework: 
\paragraph{Tracking of desired distance from the object}
Figure \ref{inplace}(a) shows the distance between the humanoid's COM and the box converging to the desired 0.6 m over time, demonstrating that the MPC generates footstep patterns that keep the robot close to the box.
Stiffness modulation further enhances this behavior by enabling adaptive stepping to achieve convergence to desired distances. The small oscillations observed in the trajectory are primarily caused by the impacts generated during the robot's stance foot switching in in-place walking.
\paragraph{Evaluation of load sharing}
Figure \ref{inplace}(b) shows the z-direction forces applied by the humanoid and the human during the task. The humanoid carries more than half of the load, thereby reducing the vertical effort required from the human by more than 50\%. The CoM of the box is located closer to the human-side interaction ports than to those of the robot. Consequently, as the robot enforces a zero-pitch objective for the box, it ends up supporting more than 50\% of the load. For collaborative tasks, particularly those difficult for a human to perform alone, such as heavy object transportation or long-duration efforts, effective load sharing reduces human effort, enhances stability, and promotes ergonomics, thereby improving the overall efficiency of the task. 
\paragraph{Evaluation of stability through footstep patterns}  
The concept of a capture point is widely used for controlling legged robots and also serves as a metric for stability. Boundedness of the capture point about the stance foot indicates stability in the sense of balance. The farther the capture point from the stance foot, the more prone it is to falling. To evaluate stability in the presence of external forces, we use a modification of this capture point to account for the external forces and their effect on the stability of the humanoid. The mathematical definitions of capture point emerge from the divergent component of the LIP model. The dynamics of this model in the presence of external forces is given by:
\begin{align}
    m_\mathrm{c}\ddot{\mathbf{x}}^\mathrm{c} = m_\mathrm{c}\omega^2_o(\mathbf{x}^\mathrm{c} - \mathbf{u}) + \mathbf{F}_{ext}
\end{align}

Here, $\mathbf{F}_{ext}$ is the equivalent force at the CoM that produces the same moment about the stance foot as the interaction forces at the hands. The above equation can be rearranged as:
\begin{align}
    \ddot{\mathbf{x}}^\mathrm{c} = \omega^2_o\big(\mathbf{x}^\mathrm{c} + \frac{\mathbf{F_{ext}}}{m_\mathrm{c}\omega^2_o}- \mathbf{u}\big). 
    \label{eq:LIPExternalForce}
\end{align}
We define a force and position combined state variable ($\boldsymbol{\gamma}^c)$ as:
\begin{align}
    \boldsymbol{\gamma}^\mathrm{c} = \mathbf{x}^\mathrm{c} + \frac{\mathbf{F}_{ext}}{m_\mathrm{c}\omega^2_o}
    \label{eq:ModifiedCP}
\end{align}
Assuming the derivatives of forces ($\mathbf{F}_\mathrm{ext})$ are zero and substituting $\boldsymbol{\gamma}^c$ in ~\eqref{eq:LIPExternalForce} we get:
\begin{align}
    \ddot{\boldsymbol{\gamma}}^\mathrm{c} = \omega^2_o(\boldsymbol{\gamma}^\mathrm{c}-\mathbf{u}).
\end{align}
These dynamics are similar to the dynamics of the LIP model. Using concepts from the capturability frameworks \cite{Koolen2012Capturability}, we define the modified capture point, $\boldsymbol{\xi}^\mathrm{c}$ as:
\begin{align}
    \boldsymbol{\xi}^\mathrm{c} = \boldsymbol{\gamma}^c + \frac{\dot{\boldsymbol{\gamma}}}{\omega_o}
\end{align}
Walking stability can be ensured if the modified capture point is bounded with respect to the stance foot. For all experiments, we compute the \emph{modified capture point offset} denoted by $\boldsymbol{\epsilon}^\mathrm{c}$ and defined as the relative position of the modified capture point from the stance foot, $\boldsymbol{\xi}^\mathrm{c}-\mathbf{u}$ to demonstrate stability.

Figure \ref{inplace}(c,d) shows the plots of this offset in the forward and lateral directions obtained by the transformation $\big(\mathbf{R}(\theta^\mathrm{s})\big)^\top\boldsymbol{\epsilon}^\mathrm{c}$. The results demonstrate that the offset exhibits a near-periodic and bounded behavior, which indicates stability/balance. Importantly, it does not diverge, confirming that the MPC generates stepping patterns that maintain stability even in the presence of external interaction forces, thereby preventing falls.

\subsection{Evaluation of the Effect of Compliance}
There are two primary ways in which compliant behavior can be realized in human–humanoid collaboration: through interactions at the hands and through the stepping pattern. Prior work \cite{Kumbhar2025MPC} has demonstrated the use of an MPC framework to generate stepping patterns that track a desired compliant response. In this work, we extend these capabilities by also leveraging the humanoid’s hands to actively contribute to compliance during co-transportation. Our framework enables the hands and the stepping pattern to execute desired trajectories coming from different levels of compliance given by the stiffness and damping parameters.

We hypothesize that collaboration is most efficient when the stepping pattern follows a relatively stiff trajectory, while the hands exhibit compliant behavior.
The compliance at the hands effectively creates a cushioning layer between the humanoid and the box, which not only promotes more stable locomotion but also improves the human’s efficiency in maneuvering the object.
On the other hand, prescribing a low-compliance (i.e., stiff) trajectory for the stepping pattern enhances the robot’s reactivity, thereby increasing human efficiency in moving the box. However, when the locomotion speeds are too high or the capture point drifts far from the stance foot, the risk of falling is higher for the case of stiff trajectories.  
\begin{table}[t!]
    \centering
    \caption{Average efficiency ($\bar{\eta}$) for the four cases during the periodic motion task.}
    \label{efficiency_table}
    \renewcommand{\arraystretch}{1.2}
    \setlength{\tabcolsep}{10pt}
    \begin{tabular}{ccc}
        \hline
        \textbf{Case} & \textbf{Parameters(N/m \& Ns/m)} & $\bar{\eta}$ \\
        \hline
        \hline
        \vspace{0.1cm}
        1 & 
        \begin{tabular}{@{}c@{}} 
            $K^x_a = 500$, $B^x_a = 40$ (Low) \\ 
            $K^x_h = 500$, $B^x_h = 40$ (Low) 
        \end{tabular} & 
        0.818 \\
        \hline
        \vspace{0.1cm}
        2 & 
        \begin{tabular}{@{}c@{}} 
            $K^x_a = 25$, $B^x_a = 10$ (High) \\ 
            $K^x_h = 25$, $B^x_h = 10$ (High) 
        \end{tabular} & 
        0.677 \\
        \hline
        \vspace{0.1cm}
        3 & 
        \begin{tabular}{@{}c@{}} 
            $K^x_a = 500$, $B^x_a = 40$ (Low) \\ 
            $K^x_h = 25$, $B^x_h = 10$ (High) 
        \end{tabular} & 
        0.814 \\
        \hline
        \vspace{0.1cm}
        4 & 
        \begin{tabular}{@{}c@{}} 
            $K^x_a = 25$, $B^x_a = 10$ (High) \\ 
            $K^x_h = 500$, $B^x_h = 40$ (Low) 
        \end{tabular} & 
        0.524 \\
        \hline
        \vspace{0.1cm}
    \end{tabular}
\end{table}

We prove our hypothesis through a dynamic experiment, where the human as a leader is asked to move the box forward and backward in an approximately periodic fashion as shown in Fig. \ref{periodic}(top). This ensures that the position and velocity of the box change frequently for the compliance properties to show effect. We run four different experiments for all high-compliance and low-compliance combinations of the desired trajectories in the high-level planner and the control objective for hands in the low-level controller. Table \ref{efficiency_table} lists the four cases with their respective parameters. For the following sections, parameters ($\mathbf{K}_i = 25N/m, \mathbf{B}_i=10Ns/m ~\forall~ i\in\{\mathrm{h},\mathrm{a}\}$) correspond to high compliance and ($\mathbf{K}_i = 500N/m, \mathbf{B}_i=40Ns/m ~\forall~ i\in\{\mathrm{h},\mathrm{a}\}$) correspond to low compliance. Figure \ref{periodic}(a) shows the locally forward box velocity for all four combinations. All four plots have almost the same amplitude and frequency of oscillation, showing a similar intended motion profile of the human in all four cases of compliance.

To compare the four cases of compliance, we introduce a collaboration efficiency metric. This metric is defined as the ratio between the net effort applied to the box to accomplish the task and the total effort exerted by both the human and the humanoid on the box. Effort is quantified as the time integral of absolute power, capturing the total work performed irrespective of whether it is assistive or resistive. The net effort $E_\mathrm{n}$ is given by:
\begin{align}
E_\mathrm{n} &=  \int_{0}^{\mathrm{w}} |(\mathbf{F}_\mathrm{h}+\mathbf{F}_\mathrm{r})\cdot\dot{\mathbf{x}}^\mathrm{b}| \,dx  
\label{eq:E_net}
\end{align}
Here, $\mathbf{F}_\mathrm{h}$ and $\mathbf{F}_\mathrm{r}$ are the net forces applied by the human and the robot on the box, while $\mathrm{w}$ is the time window of the integral, which allows us to quantify how the effort changes over time.
The total effort ($E_\mathrm{s}$) is defined as the sum of the individual efforts of the human and the humanoid and it is given by:
\begin{align}
    E_\mathrm{s} &=  \int_{0}^{\mathrm{w}} (|\mathbf{F}_\mathrm{h}\cdot\dot{\mathbf{x}}^\mathrm{b}|+|\mathbf{F}_\mathrm{r}\cdot\dot{\mathbf{x}}^\mathrm{b}|) \,dx  
    \label{eq:E_total}
\end{align}
Let $\eta$ be the efficiency of the collaboration defined as the ratio of net effort to the total effort:
\begin{align}
    \eta = \frac{E_\mathrm{n}}{E_\mathrm{s}}
    \label{eq:efficiency}
\end{align}

The proposed metric quantifies the efficiency with which net work is transferred to the object during the collaborative task. A lower efficiency indicates that a significant portion of the individual efforts are either dissipated or cancel each other out, i.e., the agents are working in opposition. This reduces the net work performed on the object. 
This metric reflects not only task-level performance but also serves as a proxy for the degree of cooperation between the human and the humanoid. Figure \ref{periodic}(b) shows the efficiency calculated over time for the four cases, using a sliding window ($\mathrm{w}$) of 7.67 s and a stride of 15.34 ms
to smooth out high frequencies.
As can be seen, the efficiency for Cases 1 and 3, corresponding to a low compliance objective for the stepping pattern, is higher than that of the other cases. We define the mean efficiency, $\bar{\eta}$, as the average value of the efficiency computed using~\eqref{eq:efficiency} across all sliding windows over the duration of the experiment. Table \ref{efficiency_table} shows the mean efficiencies ($\bar{\eta}$) for the four cases. Cases 1 and 3 have the highest and comparable efficiencies, supporting our hypothesis that low-compliance footstep objectives play a critical role in enhancing efficiency during collaborative tasks.

In all cases, the humanoid is lifting more than half of the load of the box, as shown in Fig. \ref{periodic}(c,d) by the vertical forces applied by the human and the humanoid. Again, higher load sharing contribution of the humanoid is a consequence of the robot's zero box-pitch objective. Significantly reduced human load enables the co-transportation of heavy objects and the execution of long-duration tasks. Figure \ref{periodic}(e,f) shows the modified capture point offsets for the local $X'$ or forward axis, ${\epsilon}^\mathrm{c}_{x,l}$, and $Y'$ or lateral axis, ${\epsilon}^\mathrm{c}_{y,l}$. As can be seen, the offset stays bounded in both directions, indicating that the robot is not falling and is able to generate a stepping pattern to maintain balance. Additionally, for the lateral direction, it is similar or less in amplitude compared to the in-place walking task, indicating less correlation between the motion in the forward and lateral directions. Perturbations in the forward direction do not affect stability in the lateral direction. Figures \ref{periodic}(g,h) show the forward-direction forces exerted on the box by the human and the humanoid. The framework demonstrates robustness by maintaining balance even under external forces exceeding 100N, with peak values approaching 150N. 


\subsection{Straight Line Walking Evaluation}
Using this framework, we were able to achieve varying forward speeds for the box carried by the human-humanoid pair, with a maximum of approximately 0.7 m/s.
Figure \ref{speeds} shows the different speeds achieved by the box during the collaboration.  Under those conditions, we analyzed the efficiency and robot behavior under high-compliance ($K^x_\mathrm{h} = 25.0 N/m, B^x_\mathrm{h} = 10.0Ns/m$) and low-compliance ($K^x_\mathrm{h} = 500.0 N/m, B^x_\mathrm{h} = 40.0Ns/m$) objectives for the hands, while keeping the footstep pattern objective fixed, to highlight the contribution of the hands to efficient collaboration. Figure \ref{straight}(a) shows the locally forward box velocity for both cases. Both plots have almost the same trajectory, showing a similar intended motion profile of the box in both cases of compliance.

For effort calculations in the efficiency metric, we consider only forward-direction forces and velocities, reflecting the predominantly straight walking nature of the task. Specifically, the dot product computed in \eqref{eq:E_net} and  \eqref{eq:E_total} converts to a scalar product between the local $x$ or forward axis force measurements($F^x_\mathrm{h}$ and $F^x_\mathrm{r})$ and the box velocity ($\dot{x}^\mathrm{b})$.  

Figure \ref{straight}(b) shows the efficiency calculated over time for both cases.
The efficiency for the case of high compliance is consistently higher than that of low compliance.
This implies that high hand compliance enables more efficient collaboration, with the dyad working in synchrony to accomplish the task.
Figure \ref{straight}(c,d) shows the forces applied by the human and the humanoid in the vertical direction on the box. Similar to previous experiments, the humanoid contributes to more than 50\% of the load, significantly reducing human effort. 
Figure \ref{straight}(e,f) shows the evolution of the modified capture point offset, which remains bounded in both the forward and the lateral directions without diverging. Figure \ref{straight}(g,h) shows the forces applied by the human and the humanoid on the box in the local forward direction. These forces remain minimal, indicating that the box is easily manipulated by the human.

\subsection{Complex Maneuvering Evaluation}


The proposed framework enables the humanoid to execute complex maneuvers, including lateral walking and turning. In the pure translation task shown in Fig. \ref{complex} (left), the robot sequentially moves forward, laterally left, laterally right, and finally backward to its original position, passively following the human's lead. Figure \ref{complex} (right) illustrates a semicircular task that combines translation and yaw rotation, where the human moves the box along a semicircular path and the robot follows passively to complete the motion. It is important to note that in all tasks, the humanoid has no prior knowledge of the human’s intended trajectory; it simply responds by following the motion of the object.

\section{CONCLUSION}

We presented an MPC–QP control framework with stiffness modulation to enable physical human–humanoid collaboration in co-transportation tasks. The framework incorporates a high-level planning layer based on an admittance and an I-LIP driven MPC to generate compliant motion plans, which are executed on the humanoid Digit through a low-level QP that accounts for both robot and object dynamics. Stiffness modulation was introduced to adaptively regulate coupling between the robot and the object, ensuring convergence to the desired relative configuration while maintaining flexibility in interaction.

The proposed approach was validated through real-world experiments on Digit, where we introduced an efficiency metric to jointly assess task performance and inter-agent coordination. Experimental results demonstrated that compliance plays a critical role in effective collaboration, improving efficiency and shaping desirable trajectory characteristics across different levels of control. These results show that humanoid robots can achieve a wide range of co-transportation behaviors, including forward motion, lateral maneuvers, and turning, by passively adapting to human intent without requiring prior knowledge of the trajectory.

This work makes three main contributions: (i) a hierarchical control framework that integrates stiffness modulation into MPC–QP layers for human–humanoid co-transportation, (ii) an efficiency metric for quantitatively evaluating collaboration, and (iii) an experimental demonstration on a full-scale humanoid robot validating the effectiveness of compliance in enhancing dyadic transport and showcasing the robot's capability to perform diverse and complex co-transportation tasks.

The significance of this problem lies in enabling humanoid robots to perform complex, coordinated tasks alongside humans in unstructured, human-centered environments. Such capabilities are essential for a wide range of real-world applications, including collaborative transport in warehouses and assembly lines, patient handling in healthcare, and everyday assistance in homes. By adapting seamlessly to human motions, such as turning, lateral shifts, and variable speeds, humanoids equipped with the proposed framework become versatile and effective collaborators, paving the way for practical deployment in dynamic real-world settings.



%

\appendices




\ifCLASSOPTIONcaptionsoff
  \newpage
\fi

\balance

\bibliographystyle{IEEEtran}
 
\bibliography{IEEEabrv, ref}






\end{document}